\documentclass[10pt,journal,compsoc]{IEEEtran}

\usepackage{color}
\usepackage{caption}
\usepackage{tcolorbox}
\usepackage{times}
\usepackage{helvet}
\usepackage{courier}
\usepackage{amsmath}
\usepackage{amssymb}
\usepackage{amsthm}
\usepackage{graphicx}
\usepackage{array}
\usepackage{booktabs}
\usepackage{amsopn}
\usepackage{amsmath,bm}
\usepackage{algorithm}
\usepackage{algorithmic}
\usepackage{enumerate}
\usepackage{color}
\usepackage{multirow}
\usepackage{bbding}
\usepackage{threeparttable}
\usepackage{threeparttable} 
\usepackage{dsfont} 
\usepackage{makecell} 
\usepackage{url}            
\usepackage{CJKutf8}
\usepackage{subcaption}

\newcommand{\eg}{\emph{e.g.}}
\newcommand{\etal}{\emph{et al.}}
\newcommand{\ie}{\emph{i.e.}}

\newcommand{\wrt}{{\emph{w.r.t. }}}
\newcommand{\aka}{\emph{a.k.a }}

\newtheorem{theorem}{Theorem}
\newtheorem{lemma}{Lemma}

\newcommand{\qileft}{[\kern-0.15em[}
\newcommand{\qiLeft}{\left[\kern-0.4em\left[}
\newcommand{\qiright}{]\kern-0.15em]}
\newcommand{\qiRight}{\right]\kern-0.4em\right]}

\newcommand{\R}{\mathbb{R}}

\newcommand{\T}{\mathcal{T}}

\newcommand{\MSA}{{\rm MSA}}
\newcommand{\MLP}{{\rm MLP}}
\newcommand{\Concat}{{\rm Concat}}

\graphicspath{{figs/}}
\begin{document}
	\begin{CJK}{UTF8}{gbsn}
		\title{PanGu-$\pi$: Enhancing Language Model Architectures via Nonlinearity Compensation }
		\author{Yunhe Wang, Hanting Chen, Yehui Tang, Tianyu Guo, Kai Han, Ying Nie, Xutao Wang, Hailin Hu, Zheyuan Bai, Yun Wang, Fangcheng Liu, Zhicheng Liu, Jianyuan Guo, Sinan Zeng, Yinchen Zhang, Qinghua Xu, Qun Liu, Jun Yao, Chao Xu, and Dacheng Tao~\IEEEmembership{Fellow,~IEEE}
			\IEEEcompsocitemizethanks{
				\IEEEcompsocthanksitem Corresponding to Yunhe Wang (Huawei Noah's Ark Lab). E-mail: yunhe.wang@huawei.com.
				\IEEEcompsocthanksitem Acknowledgments: This work is jointly funded by Huawei 2012 Labs and Huawei Group Finance. We also thank the work from both the data engineering team and IT architecture team.
			}
		}
	\markboth{ }
	{Wang \MakeLowercase{\textit{et al.}}: PanGu-$\pi$: Enhancing Language Model Architectures via Nonlinearity Compensation}
	
	\IEEEtitleabstractindextext{%
		\begin{abstract}
			The recent trend of large language models (LLMs) is to increase the scale of both model size (\aka the number of parameters) and dataset to achieve better generative ability, which is definitely proved by a lot of work such as the famous GPT and Llama.
			However, large models often involve massive computational costs, and practical applications cannot afford such high prices.
			However, the method of constructing a strong model architecture for LLMs  is rarely discussed.
			We first analyze the state-of-the-art language model architectures and observe the feature collapse problem. Based on the theoretical analysis, we propose that the nonlinearity is also very important for language models, which is usually studied in convolutional neural networks for vision tasks. 
			The series informed activation function is then introduced with tiny calculations that can be ignored, and an augmented shortcut is further used to enhance the model nonlinearity.
			We then demonstrate that the proposed approach is significantly effective for enhancing the model nonlinearity through carefully designed ablations; thus, we present a new efficient model architecture for establishing modern, namely, PanGu-$\pi$. 
			Experiments are then conducted using the same dataset and training strategy to compare PanGu-$\pi$ with state-of-the-art LLMs. 
			The results show that PanGu-$\pi$-7B can achieve a comparable performance to that of benchmarks with about 10\% inference speed-up, and PanGu-$\pi$-1B can achieve state-of-the-art performance in terms of accuracy and efficiency.
			In addition, we have deployed PanGu-$\pi$-7B in the high-value domains of finance and law, developing an LLM named YunShan for practical application. The results show that YunShan can surpass other models with similar scales on benchmarks.
		\end{abstract}
		
		}
	
	\maketitle
	
	\IEEEdisplaynontitleabstractindextext
	
	\IEEEpeerreviewmaketitle
	
	\IEEEraisesectionheading{\section{Introduction}\label{Sec:Introduction}}
\IEEEPARstart{L}{arge} language models (LLMs) have significantly evolved and are capable of a wide range of NLP tasks such as machine translation, text summarization, and dialogue. Following the scaling law~\cite{kaplan2020scaling}, a series of studies confirm significantly improved performances and emergent abilities~\cite{wei2022emergent} on downstream tasks by scaling up the model size and the data size, \eg, GPT-3 with 175B parameters~\cite{gpt3} and PaLM with 540B parameters~\cite{palm}. Recently, a remarkable innovation, ChatGPT, was introduced with the ability to interact with humans in a conversational way. The success of ChatGPT is attributed to pre-training on extensive textual data and fine-tuning with alignment to human preferences. This paradigm has a profound impact on subsequent research and real-world applications~\cite{gpt4,xi2023rise}. The emergence of ChatGPT has inspired the community to develop more excellent large models, including LLaMA~\cite{llama}, ChatGLM~\cite{zeng2023glm-130b}, and Baichuan~\cite{yang2023baichuan}, which in turn drive the vigorous development of the LLM field.

\begin{figure*}[t]
	\centering
	\includegraphics[width=1.0\linewidth]{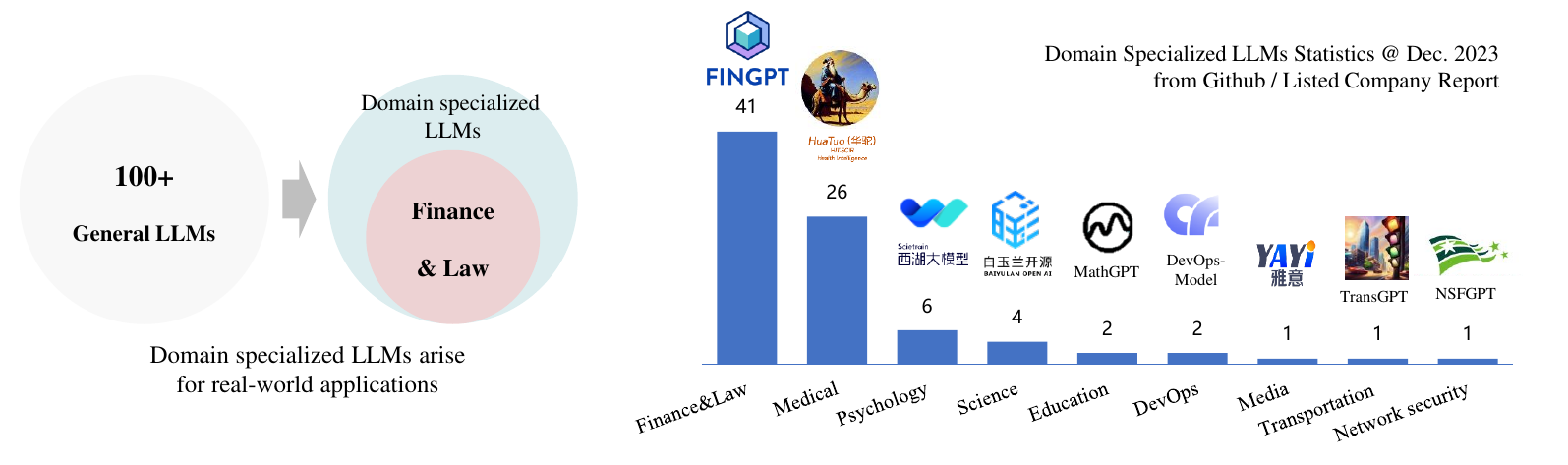}
	\caption{Statistics of domain specialized LLMs. The general LLMs face challenges in supporting industry applications, leading to a growing emphasis on domain specialized LLMs. Among them, the fields of finance and law are particularly active.}
	\label{fig:domain-llm}
\end{figure*}

In addition to general-purpose LLMs, high-value domain-specific large models are also being extensively researched to promote the implementation and practical application of LLMs. For example, LaWGPT~\cite{LaWGPT} enhances the fundamental semantic understanding in the field of law with a specialized legal vocabulary and extensive pre-training on a large-scale Chinese legal corpus. FinGPT~\cite{fingpt} develops an open-source LLM in a data-centric approach. Huatuo~\cite{wang2023huatuo} builds a Chinese medical instruction fine-tuning dataset and enhances its performance in question-answering within the medical field. As shown in Figure~\ref{fig:domain-llm}, our analysis of the industry distribution of domain-specialized LLMs reveals that those in the finance and law domains attract the most attention due to their widespread demand and commercial value.

The Transformer model introduced in 2017~\cite{vaswani2017attention} is the foundational architecture for many LLMs. The core components of the Transformer model include the multi-head self-attention (MSA) and the feed-forward network (FFN). MSA computes attention scores for each token in the input sequence with respect to all other tokens, capturing relationships and dependencies. FFN is performed on each token separately which provides more nonlinear transformation and model capacity. Transformer architectures are utilized to build encoder (\eg, BERT) and decoder (\eg, GPT-2) for NLP tasks. In LLM, decoder-only architecture is widely used by predicting the next token with the context information~\cite{gpt3,gpt4}. Beyond the standard Transformer architecture, a series of studies have explored architecture modification (especially MSA~\cite{rwkv,katharopoulos2020transformers} or FFN~\cite{palm,llama}) seeking better performance. PaLM~\cite{palm} and LLaMA~\cite{llama} use SwiGLU-based FFN~\cite{swiglu} which consists of the component-wise product of two linear layers, showing significantly increased generation quality. RWKV (Receptance Weighted Key Value)~\cite{rwkv} proposes an RNN-style attention mechanism to alleviate the quadratic complexity in standard MSA. Switch Transformer~\cite{fedus2022switch} allocates different parameters for each input example and results in a sparsely-activated model.

The development of an excellent LLM is a complex system engineering, which includes data preparation, data cleaning, model architecture, cluster commutation, and optimizer. The model architecture design is one of the most important components and determines the maximum performance potential of the deployed LLM. Among the recent projects in 2022-2023, the popular versions that are often used for secondary development are GPT-3~\cite{gpt3} and LLaMA~\cite{llama}. By inheriting the observations and analysis of our previous work~\cite{augvit}, we find that the feature collapse problem also affects the expressive power of these well-designed Transformer architectures. Taking LLaMA as an example, we empirically analyze its feature collapse phenomenon using the rank metric~\cite{dong2021attention}. The feature rank diminishes significantly in deeper layers, leading to a greater similarity among all tokens. This greatly degrades the generation quality and diversity of LLMs. We also theoretically analyze the feature collapse problem in Transformer architecture (details in Section~\ref{lilun}). Through theoretical analysis, we have discovered that nonlinearity significantly impacts the capabilities of the Transformer model. Enhancing nonlinearity can effectively mitigate the issue of feature collapse and improve the expressive power of the Transformer model. We intend to construct stronger LLM architectures by approaching them from a nonlinear perspective.

\begin{figure*}[t]
	\centering
	\includegraphics[width=1.0\linewidth]{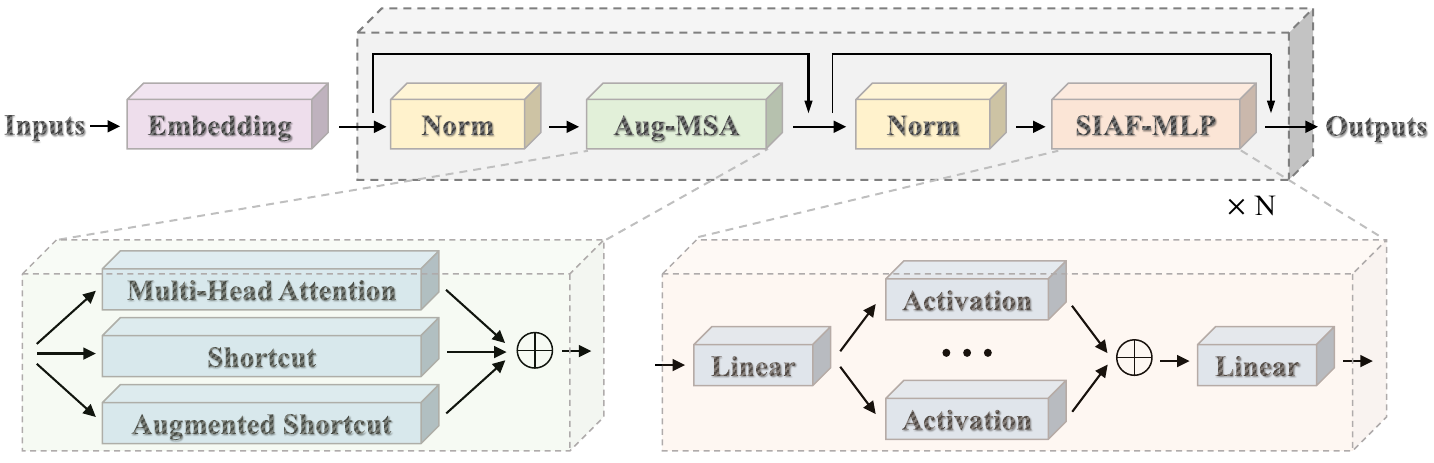}
	\caption{The diagram of the proposed PanGu-$\pi$ architecture. The series activation function is adapted to FFN, and the augmented shortcuts are integrated into MSA, which effectively introduces more nonlinearity into the Transformer architecture.}
	\label{fig}
\end{figure*}

In this paper, we introduce a new architecture for LLMs to address the feature collapse problem via nonlinearity compensation, named PanGu-$\pi$. We introduce more nonlinearity from two approaches in both FFN and MSA modules without significant increasing the model complexity. First, the series-based activation function with multiple learnable affine transformation is equipped in FFN, which can effectively enhance the nonlinearity of the entire network with negligible calculations. Then, the augmented shortcut is paralleled with the main branch of each MSA module to eschew the rank collapse. To maintain the model efficiency, we carefully refine the augmented shortcut operation with hardware-friendly operations. The enhanced PanGu-$\pi$ architectures (see Figure~\ref{fig}) are constructed with both the series activation-based FFN and shortcut-augmented MSA. We also prove that the superposition of these two operations can enhance nonlinear compensation. We build two versions of PanGu-$\pi$ with different model sizes, \ie, PanGu-$\pi$-7B and PanGu-$\pi$-1B. By training on a large-scale corpus, our PanGu-$\pi$ models obtain general language ability on downstream tasks. Through carefully designed ablations, we demonstrate that the proposed approach can effectively enhance the model nonlinearity and alleviate feature collapse. Thus, with the same scale of parameters, we can achieve a substantial efficiency gain via the two new modules. Extensive experiments on various NLP tasks are evaluated to compare with state-of-the-art LLMs. In a scenario with similar model size, PanGu-$\pi$ models can achieve better performance in terms of both accuracy and efficiency. In addition to the foundational abilities, we have deployed PanGu-$\pi$-7B in the high-value domains of finance and law, developing a specialized LLM named YunShan for practical application. Extensive evaluations of finance and law benchmarks also show that YunShan surpasses other state-of-the-art models with similar scales.

This work introduces a new LLM network architecture (\ie, PanGu-$\pi$) with extensive experiments and theoretical analysis with some of the ideas borrowed from our preliminary works published on NeurIPS 2023~\cite{vanillanet} and NeurIPS 2021~\cite{augvit}. This present work makes the following significant contributions. First, the previous two papers, one involving the construction of a CNN backbone~\cite{vanillanet} and the other focusing on vision Transformers~\cite{augvit}, have laid a foundation that this paper seeks to extend within LLMs. We achieved commendable results in this endeavor, and extensive experiments have validated the effectiveness of our methods. Second, the previous two papers analyzed the network design subject from distinct perspectives. In our current study, the two works are theoretically integrated and essentially address the same critical issue from a unified standpoint in the LLM domain. Third, we organically adapted series activation to FFN and integrated augmented shortcuts into MSA. These two components are complementary to each other and effectively introduce more nonlinearity into the Transformer architecture. Fourth, we developed the PanGu-$\pi$ foundation models by large-scale training and fine-tuning (SFT), which achieved state-of-the-art results in general NLP tasks for a similar model size. In addition, we extend PanGu-$\pi$ to finance and legal domains by transfer learning and obtain excellent performance on these downstream tasks.

The rest of this paper is organized as follows. Section 2 reviews related work in the field of Transformer architectures for building LLMs and the related hotspot applications. Section 3 provides a theoretical analysis of the feature collapse problem and the nonlinear capabilities of existing Transformer architectures. Section 4 introduces a nonlinear enhancement strategy based on the series activation function and augmented shortcut. Section 5 details the data, training strategies, and experimental results of the PanGu-π architecture with two models with important parameter scales, \ie, PanGu-$\pi$-7B and PanGu-$\pi$-1B. In Section 6, the PanGu-$\pi$  architecture is deployed in the high-value domains of finance and law, developing the YunShan LLM for practical application. Extensive evaluations of finance and law benchmarks also show that YunShan surpasses other state-of-the-art models with similar scales.  Section 7 concludes the entire paper and discusses future works.

\section{Related Works}

In this section, we first summarize the recent representative works in the field of LLMs. We then review the classical works for enhancing Transformer architectures. Lastly, we investigate the domain-specific large language models, especially in finance and law.

\subsection{LLMs}

With the emergence of ChatGPT~\cite{OpenAI-blog-2022-ChatGPT} from OpenAI, LLMs with billions of parameters achieved astounding performance on various natural language processing tasks. Subsequently, the latest GPT-4~\cite{OpenAI-OpenAI-2023-GPT-4} pushed the generalization capabilities of LLMs to a new level. However, the proliferation of GPT-series models is accompanied by a strict commercial orientation that is not conducive to a thriving open source community. The representative work of democratic LLMs is LLaMA~\cite{touvron2023llama}, a collection of open-source foundation language models ranging from 7B to 65B parameters. Later, a more elaborate model LLaMA2~\cite{touvron2023llama2} is introduced, appearing to be on par with some closed-source models~\cite{OpenAI-blog-2022-ChatGPT} based on human evaluations. Since its release, LLaMA has attracted extensive attention from the academia and industry. Subsequent efforts have been based on LLaMA by either instruction tuning or continual pre-training. Stanford Alpaca~\cite{taori2023alpaca} is the first LLaMA-based chatbot fine-tuned with 52K instruction-following samples generated by the self-instruct method~\cite{wang2022self}. Vicuna~\cite{chiang2023vicuna} also fine-tunes LLaMA with user-shared conversations collected from ShareGPT~\cite{Sharegpt}. In addition to the language models geared toward English-dominant applications, multilingual language models are also thriving. InternLM~\cite{team2023internlm} presents a multilingual foundational language model pre-trained on a large corpus with 1.6T tokens with a multi-phase progressive process. Baichuan2~\cite{yang2023baichuan} introduces a series of large-scale multilingual language models containing 7 billion and 13 billion parameters. PanGu-$\Sigma$~\cite{ren2023pangu} extends the dense Transformer model to a sparse one with Random Routed Experts. Qwen~\cite{bai2023qwen} introduces a comprehensive language model series that encompasses distinct models with varying parameter counts. Skywork~\cite{wei2023skywork} presents 13B LLMs trained on a corpus drawn from both English and Chinese texts with a two-stage training methodology.

\subsection{Enhanced Transformer Architectures}
While Transformer architectures have gained significant prominence in LLMs recently, there continues to be a surge of interest in their effective utilization in diverse domains, including computer vision tasks. In light of this, we review classical works dedicated to enhancing Transformer structures, with a particular focus on augmenting model nonlinearity and efficiency.

\textbf{Natural language processing domain.}
The conventional self-attention mechanism, with quadratic computational complexity, poses challenges for handling long input sequences during training and inference. To mitigate this, various structural priors on attention, including \emph{sparsity}~\cite{zaheer2020big,child2019generating,roy2021efficient,rae2019compressive,xiao2023efficient} and \emph{linear attention}~\cite{katharopoulos2020transformers,choromanski2020rethinking}, have been proposed. Notably, Reformer~\cite{kitaev2020reformer} employs locality-sensitive hashing to approximate full attention. Longformer~\cite{beltagy2020longformer} integrates local windowed attention with task-motivated global attention. Models such as GPT-3\cite{brown2020language} incorporate locally banded sparse attention methods, such as Factorized Attention~\cite{child2019generating}. There are also works focusing on replacing the attention module by incorporating recurrent models~\cite{dai2019transformer,martins2021infty,sun2023retentive}.
Hyena~\cite{poli2023hyena} trained a recurrence of gating units and implicitly parametrized long convolutions, which serves as an attention-free drop-in replacement for the traditional Transformer architecture. RWKV~\cite{peng2023rwkv} replaced the quadratic QK attention with a scalar formulation that has linear cost. RetNet~\cite{sun2023retentive} theoretically derived the connection between recurrence and attention and proposed the retention mechanism for sequence modeling. There are also efficient enhancements focused on the Feed-Forward Network (FFN). Mixture-of-Experts (MoE)~\cite{du2022glam,fedus2022switch,roller2021hash,chi2022representation,lewis2021base} has demonstrated effectiveness in the pre-training of LLMs. In addition to MoE, PaLM~\cite{chowdhery2022palm} and LLaMA~\cite{touvron2023llama} leverage the SwiGLU activation for original FFN intermediate activations. This choice is grounded in the observation that SwiGLU activations, as demonstrated in compute-equivalent experiments~\cite{shazeer2020glu}, substantially enhance quality compared to standard activation functions like ReLU, GeLU, or Swish.

\textbf{Computer vision domain.}
PVT~\cite{wang2021pyramid} and Swin~\cite{liu2021swin} utilize hierarchical structures across multiple stages, overcoming challenges posed by the original isotropic ViT~\cite{dosovitskiy2020image} for diverse computer vision tasks. Ongoing research focuses on refining local information processing~\cite{guo2022cmt,heo2021rethinking,pan2021scalable,chen2021crossvit,graham2021levit,mehta2021mobilevit,han2021transformer}, simplifying attention mechanisms~\cite{parmar2018image,liu2023efficientvit,xiong2021nystromformer}, and exploring alternative modules~\cite{guo2022beyond,guo2022hire,tang2022image,yu2022metaformer,lian2021mlp,chen2021cyclemlp}.
For example, T2T-ViT~\cite{yuan2021tokens} reduces token length through iterative aggregation, whereas TNT~\cite{han2021transformer} captures local information by dividing ViT's patches. Swin~\cite{liu2021swin} and Cswin~\cite{dong2022cswin} introduce local attention within a window and shifted window partitioning for cross-window connections. GFnet~\cite{rao2021global} employs Fast Fourier Transform for token mixing. Architectures like ResMLP~\cite{touvron2022resmlp} and MLP-Mixer~\cite{tolstikhin2021mlp}, solely rely on multi-layer perceptrons (MLPs), excluding convolutions or self-attention mechanisms.

\subsection{LLMs for Finance and Law}
In addition to the LLMs towards general purpose, the domain-specific models that are more capable of generating applied value are receiving increasing attention, with finance and law being the most representative.

\textbf{Financial LLMs. } Wu~\etal propose the first proprietary LLM with 50 billion parameters specialized for the financial domain, \ie BloombergGPT~\cite{wu2023bloomberggpt}, which is a decoder-only causal language model based on BLOOM~\cite{workshop2022bloom}. The proposed training strategy of mixing domain-specific and general-purpose data results in a balanced performance in both domains.  Unlike the proprietary BloombergGPT, FinGPT~\cite{yang2023fingpt,liu2023fingpt} takes a data-centric approach and presents an open-source LLM to researchers and practitioners. It exhibits promise in financial tasks such as sentiment classification, quantitative trading and financial fraud detection. PIXIU~\cite{xie2023pixiu} has created a large-scale multi-task instruction dataset by manually reworking open-sourced datasets~\cite{shah2022flue}. A financial LLM called FinMA is then introduced by fine-tuning LLaMA with the constructed instruction dataset. The comprehensive evaluation results including financial NLP and prediction tasks uncover the strengths and weaknesses of various LLMs when handling different financial tasks. To address the lack of open-sourced models specifically designed for Chinese finance, Zhang~\etal introduce XUANYUAN 2.0~\cite{zhang2023xuanyuan}, built upon the BLOOM~\cite{workshop2022bloom} architecture. To mitigate catastrophic forgetting, the hybrid-tuning strategy that combines the stages of pre-training and fine-tuning is proposed. By appropriately mixing the general and financial corpus in pre-training and fine-tuning, XUANYUAN 2.0 achieves impressive performance in both the general domain and financial domain. Chen~\etal propose a financial LLM DISC-FinLLM~\cite{chen2023disc} by multiple experts fine-tuning the framework based on Baichuan-13B~\cite{Baichuan13B}. Experimental results on multiple evaluation benchmarks demonstrate its promising performance.

\textbf{Legal LLMs.} The legal sector is another area that is significantly benefitting from the advancement of LLMs. BaoLuo and Lychee~\cite{BaoLuo-LawAssistant, Lychee} are lawyer assistants developed by fine-tuning Chinese legal domain QA datasets. AI Lawyer~\cite{JurisLMs} applies Active Learning to alleviate the problem of limited supervised data volume in the legal domain. FedJudge~\cite{yue2023fedjudge} focuses on the privacy of legal data and adopts Federated Learning~\cite{mcmahan2017communication} during instruction tuning, it also utilizes Continual Learning~\cite{zhou2021image} to mitigate the issue of data distribution shifts. HanFei, LaWGPT, Lawyer-llama, WisdomInterrogatory, and Fuzi.Mingcha~\cite{HanFei, LaWGPT, huang2023lawyer, wisdomInterrogatory, fuzi.mingcha} undergo a two-phase training process: further pre-training with unsupervised legal corpus to enhance the semantic understanding ability in the legal field and then supervised training with corresponding datasets. HanFei~\cite{HanFei} is the first legal LLM in China fully trained with 7B parameters and supports multi-turn dialogue, LaWGPT~\cite{LaWGPT} expands the vocabulary by adding specific legal domain terms, Lawyer-llama~\cite{huang2023lawyer} has experimented with different data permutations and training sequences during its instruction tuning phase. During inference time, LawGPT\_zh(XieZhi), LexiLaw, ChatLaw, Lawyer-llama, Fuzi.Mingcha and DISC-LawLLM~\cite{LAWGPT-zh, LexiLaw, cui2023chatlaw, huang2023lawyer, fuzi.mingcha, yue2023disclawllm} introduce a retrieval module to ensure that a definite legal document in the knowledge base supports the response. Additionally, ChatLaw~\cite{cui2023chatlaw} also involves a Keyword model for key information extraction to reduce the ambiguity of user queries. Fuzi.Mingcha~\cite{fuzi.mingcha} enables the LLM to use syllogistic reasoning to arrive at verdict predictions by training the LLM on a self-constructed dataset.

\section{Preliminaries and Motivation}\label{lilun}

In this section, we commence by dissecting the foundational architecture of the Transformer model, introducing a metric of nonlinearity to articulate its capabilities. Subsequently, we delve into an analysis of the Transformer architecture's components – the multi-head self-attention and the multi-layer perceptrons modules – scrutinizing their nonlinear expressive capability. This exploration also brings to light the limitations inherent in the current incarnations of Transformer architectures.

Recognized for the Transformer's suitability for parallel computing and the inherent complexity of its model, the Transformer has demonstrated superior precision and performance compared to the widely adopted RNN recurrent neural network. The Transformer architecture consists of two parts: the multi-head self-attention and the multi-layer perceptrons modules.

The multi-head attention module is a fundamental component of the Transformer architecture. An MSA module with $H$ heads is defined as 
\begin{equation}
	\label{eq-msa}
	\begin{aligned}
		&{\MSA} (\boldsymbol{Z}_l)=  \Concat ([\boldsymbol{A}_{lh}\boldsymbol{Z}_{l}\boldsymbol{W}^v_{lh}]_{h=1}^H)\boldsymbol{W}^o_{l},\\
		&	l \in [1,2,\cdots, L],
	\end{aligned}
\end{equation}
where $\boldsymbol{Z}_l\in \R^{N\times d}$ is the feature  of the $l$-th $\MSA$ layer,  $\boldsymbol{A}_{lh}\in \R^{N\times N}$ and $\boldsymbol{W}^v_{lh} \in \R^{d\times (d/H)}$ are the corresponding attention map and value projection matrix in the $h$-th head, respectively.  $\Concat(\cdot)$ denotes  the concatenating for features of the $H$ heads and $\boldsymbol{W}^o_l\in \R^{d\times d}$ is the output projection matrix.   
The attention matrix $\boldsymbol{A}_{lh}$ is calculated by the self-attention mechanism, \ie,
\begin{equation}
	\boldsymbol{A}_{lh}={\rm softmax}\left(\frac{(\boldsymbol{Z}_{l}\boldsymbol{W}^q_{lh})(\boldsymbol{Z}_{l}\boldsymbol{W}^k_{lh})^\top}{\sqrt d}\right),
\end{equation}
where $\boldsymbol{W}^q_{lh}\in \R^{d\times (d/H)}$ and $\boldsymbol{W}^k_{lh}\in \R^{d\times (d/H)}$ are the query and value projection matrices, respectively. Attention $\boldsymbol{A}_{lh}$ reflects the relation between different tokens, and a larger value $\boldsymbol{A}_{lh}^{ij}$ indicate that token $i$ and token $j$ have a stronger relationship. 

An MLP module is defined as
\begin{equation}
	\MLP (\boldsymbol{Z}'_{l}) = \sigma(\boldsymbol{Z}'_{l}\boldsymbol{W}'_{l_1})\boldsymbol{W}'_{l_2},~l \in [1,2,\cdots, L],
\end{equation} 

where $\boldsymbol{Z}'_{l}\in \R^{N\times d}$ is the features of the $l$-th $\MLP$ layer , $\boldsymbol{W}'_{l_1}$  and $ \boldsymbol{W}'_{l_2}\in \R^{d\times d}$ are the weight matrixs.


One of the paramount capabilities of neural networks is their nonlinear expressive capability. The higher the degree of nonlinearity, the more complex the function space the network can approximate, resulting in enhanced accuracy. In traditional Convolutional Neural Networks (CNNs), nonlinearity is primarily imparted by activation functions. However, in the Transformer architectures, the sources of nonlinearity are twofold: the self-attention mechanisms and the activation functions within the Multi-Layer Perceptrons (MLP). Hence, our analysis separately scrutinizes the nonlinear expressive capabilities of both self-attention and the MLP within the Transformer framework.

Define $\mathcal{M}_m:=\{\boldsymbol{Y}\in\mathbb{R}^{N\times m}|\boldsymbol{Y}=\boldsymbol{1x}^\top, \boldsymbol{x}^\top\in\mathbb{R}^{1\times m}\}$ as a subspace in $\mathbb{R}^{N\times m}$, where $\boldsymbol{1}=[1, 1, \dots, 1]^\top\in\mathbb{R}^{N\times1}$, $n$ is the number of tokens and $d$ is the dimension of token representation. We define the distance between matrix $\boldsymbol{H}\in\mathbb{R}^{N\times m}$
and $\mathcal{M}_m$ as $d_{\mathcal{M}_m}(\boldsymbol{H}):=\min_{\boldsymbol{Y}\in\mathcal{M}_m} \Vert
\boldsymbol{H}-\boldsymbol{Y}\Vert_F$, where $\Vert\cdot\Vert_F$ is the Frobenius norm. $d_{\mathcal{M}_m}(\boldsymbol{Z_l})$~\cite{shi2022revisiting} is a commonly used metric to measure the capability and nonlinearity of the Transformer architecture. Next, we investigate the distance between $\boldsymbol{Z_l}$ the output of layer $l$  and subspace $\mathcal{M}_d$.

We begin by examining the nonlinear expressive capabilities of the self-attention modules. The self-attention matrix within the Transformer can be intuitively likened to the normalized adjacency matrix of a corresponding graph. Viewed through a graph perspective, the self-attention layer can be seen as equivalent to a Graph Neural Network (GNN) operating on a fully connected graph with normalized edge weights. Excessive self-attention layers like GNN layers result in excessive smoothing, with node vector representations tending to be the same, resulting in convergence to a specific low-rank subspace for any given input.

The following theoretical analysis utilizes the formula of the self-attention layer to shed light on the intricacies of this phenomenon. We study how the self-attention layer converges with low rank based on matrix projection. Our analysis involves the definition of a subspace $\mathcal{M}$ characterized by a unique property in which each row vector of its elements is the same. By scrutinizing the behavior of the self-attention layer in relation to matrix projection, we aim to unravel the underlying mechanisms that drive the convergence of node vector representations toward a specific low-rank subspace. This exploration is essential for gaining deeper insights into the nuances of self-attention in Transformers, shedding light on the intricacies of their functioning and providing a theoretical foundation for potential enhancements and optimizations.

\begin{lemma} 
	\label{lm1}
	For self-attention matrix $\boldsymbol{A} \in \mathbb{R}^{N\times N}$, any weight matrix $\boldsymbol{W} \in \mathbb{R}^{d\times m}$, any $\boldsymbol{H},\boldsymbol{B}\in \mathbb{R}^{N\times d}$, $\alpha_1, \alpha_2 \geq 0$ and $\sigma$ is the nonlinear Lipschitz continuous activation function, we have:
	\begin{align*}
		d_{\mathcal{M}_m}(\boldsymbol{HW}) &\leq sd_{\mathcal{M}_d}(\boldsymbol{H}), \\
		d_{\mathcal{M}_d}(\sigma(\boldsymbol{H})) &\leq Ld_{\mathcal{M}_d}(\boldsymbol{H}), \\
		d_{\mathcal{M}_d}(\alpha_1 \boldsymbol{H}+\alpha_2 \boldsymbol{B}) &\leq \alpha_1 d_{\mathcal{M}_d}(\boldsymbol{H}) + \alpha_2 d_{\mathcal{M}_d}(\boldsymbol{B}), \\
		d_{\mathcal{M}_d}(\boldsymbol{A}\boldsymbol{H}) &\leq \sqrt{\lambda_{\max}} d_{\mathcal{M}_d}(\boldsymbol{H}) \label{eq:novel},
	\end{align*}
	where $s$ is the largest singular value of $\boldsymbol{W}$, $\lambda_{\max}$ is the
	largest eigenvalue of
	$\boldsymbol{A}^\top(\boldsymbol{I}-\boldsymbol{ee}^\top)\boldsymbol{A}$
	and $L$ is the Lipschitz constant of activation function $\sigma(\cdot)$.
\end{lemma}

Applying the lemma~\ref{lm1} to the single-head self-attention layer, we can obtain its low-rank attenuation.

\begin{theorem} 
	\label{th1-plain} 
	Given a model stacked by the MSA modules,  the diversity $d_\mathcal{M}(\boldsymbol{Z_l})$ of feature  in the $l$-th layer can be bounded by that of input data $\boldsymbol{Z_0}$, \ie, 
	\begin{align*}
		&d_{\mathcal{M}_m}(\boldsymbol{AZ}_l\boldsymbol{W})\leq \sqrt{\lambda} s\upsilon_1 d_{\mathcal{M}_d}(\boldsymbol{Z}_l). 
	\end{align*}
	where $s>0$ is the largest element of all singular values of all $\boldsymbol{W}$ and
	$\lambda$ is the largest eigenvalue of all
	$\boldsymbol{A}^\top(\boldsymbol{I}-\boldsymbol{ee}^\top)\boldsymbol{A}$
	for each self-attention matrix $\boldsymbol{A}$.
\end{theorem} 

For the low-rank matrix projection of concat matrices, we have the following lemma:

\begin{lemma} 
	\label{lm2}
	For block matrix $\boldsymbol{H}_h \in \mathbb{R}^{N\times m}$, we have:
	\begin{align*}
		d_{\mathcal{M}_{Hm}}(\Concat ([\boldsymbol{H}_h]_{h=1}^H))^2 &= \sum_{h=1}^{H}d_{\mathcal{M}_m}(\boldsymbol{H}_h)^2,
	\end{align*}
	
\end{lemma} 

Applying the theorem~\ref{th1-plain} and the lemma~\ref{lm2} to the formula~\ref{eq-msa}, we can see how the multi-headed self-attention layer decays layer by layer into the low-rank space.

\begin{theorem} 
	\label{th2-plain} 
	Given a model stacked by the MSA modules,  the diversity $d_\mathcal{M}(\boldsymbol{Z}_l)$ of feature in the $l$-th layer can be bounded by that of input data $\boldsymbol{Z}_0$, \ie, 
	\begin{align*}
		&d_{\mathcal{M}_d}(MSA(\boldsymbol{Z}_l))\leq \sqrt{\lambda H}s\upsilon_1 d_{\mathcal{M}_d}(\boldsymbol{Z}_l). \\
		&d_{\mathcal{M}_d}(\boldsymbol{Z}_l)\leq (\sqrt{\lambda H} s\upsilon_1)^{l}d_{\mathcal{M}_d}(\boldsymbol{Z}_0). 
	\end{align*}
	where $H$ is number of heads, $s>0$ is the largest element of all singular values of all $\boldsymbol{W}^v_{lh}$
	and $\upsilon_{1}$ is the largest element of all singular values of all $\boldsymbol{W}^o_{l}$.
\end{theorem} 

Assume further that $\boldsymbol{A}$ is doubly stochastic (so that $\boldsymbol{A}^\top \boldsymbol{e}=\boldsymbol{e}$) with positive entries. Then by Perron–Frobenius theorem, $\boldsymbol{A}^\top\boldsymbol{A}$ has a maximum eigenvalue $1$ with associated eigenvector $\boldsymbol{e}$ as well.
In this case, the matrix $\boldsymbol{A}^\top(\boldsymbol{I}-\boldsymbol{ee}^\top)\boldsymbol{A}=\boldsymbol{A}^\top\boldsymbol{A}-\boldsymbol{ee}^\top$ has a maximum eigenvalue $\lambda_{max}<1$.

$\sqrt{\lambda H} s\upsilon_1$ is usually smaller than 1, so the feature diversity $d_{\mathcal{M}_d}(\boldsymbol{Z}_l)$ decreases rapidly as the network depth increases. Recursively, $\boldsymbol{Z}_l$ will converge toward
subspace $\mathcal{M}_d$ if $\sqrt{\lambda H} s\upsilon_1<1$ and all representations are the same, resulting in over-smoothing. In conclusion, the nonlinear expressive capabilities of the vanilla self-attention module are limited.

We then focus on the Multi-Layer Perceptrons (MLP) of the Transformer architecture.

\begin{theorem} 
	\label{th3-mlp} 
	Given a model stacked by the MLP modules,  the diversity $d_{\mathcal{M}_d}(\boldsymbol{Z}'_l)$ of feature in the $l$-th layer can be bounded by that of input data $\boldsymbol{Z}'_0$, \ie, 
	\begin{align*}
		&d_{\mathcal{M}_d}(\MLP(\boldsymbol{Z}'_l))\leq  Ls\upsilon_2 d_{\mathcal{M}_d}(\boldsymbol{Z}'_{l}). \\
		&d_{\mathcal{M}_d}(\boldsymbol{Z}'_l)\leq (Ls\upsilon_2)^{l}d_{\mathcal{M}_d}(\boldsymbol{Z}'_0). 
	\end{align*}
	where $s>0$ is the largest element of all singular values of all $\boldsymbol{W}'_{l_1}$, $\upsilon_{2}$ is the largest element of all singular values of all $\boldsymbol{W}'_{l_2}$ and $L$ is the Lipschitz constant of activation function $\sigma(\cdot)$.
\end{theorem} 

The analysis from the prior proofs reveals that the diversity of MLP modules is constituted by two elements: the eigenvalues of parameters and the Lipschitz constant of the activation functions. In neural networks, parameters are typically normalized, which means the maximum eigenvalues of these parameters are bounded. Furthermore, the parameters in a neural network are learned through backpropagation. Given these factors, it becomes challenging to impose restrictions on the eigenvalues of parameters. Consequently, the activation functions in the MLP emerge as the most crucial aspect of their nonlinear expressive capabilities.



\section{PanGu-$\pi$ Modules and Architectures}\label{fangfa}

In this section, we first propose the series informed activation function to enhance the nonlinearity of the MLP module. Then, we introduce the augmented shortcuts to improve MSA modules in the Transformer architecture. Finally, we prove that the combination of the two techniques results in a novel and stronger Transformer model.

\subsection{Augmented Shortcut}

\begin{figure}[t]
	\centering
	\includegraphics[width=1.0\linewidth]{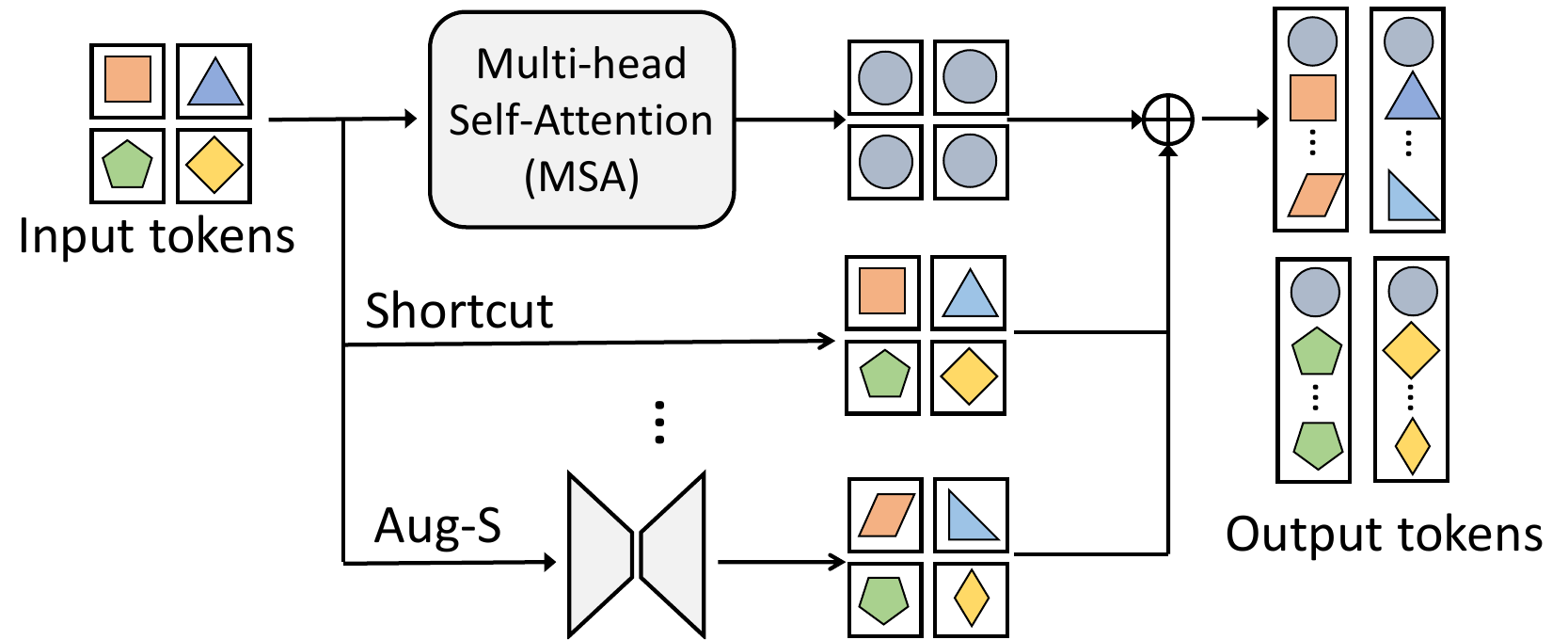}
	\caption{The diagram of MSA module equipped with augmented shortcuts, where different patterns (rectangle, triangle, etc.) denote different features from various tokens. The original identity shortcut copies the input feature while the augmented shortcuts (Aug-S) project features of each input token to diverse representations.}
	\label{fig-aug}
\end{figure}



As discussed in Section~\ref{lilun}, a pure attention suffer serve a feraure collapse  problem.  The typical LLM architecture only equips each MSA module with a single shortcut connection, which is an identity projection and directly copies the input features to the outputs. This simple formulation may not have enough representation capacity to improve the feature diversity maximally. We aim to refine the existing shortcut connections in vision Transformers and explore efficient but powerful augmented shortcuts to produce visual features with higher diversity.

We propose augmented shortcuts to alleviate the feature collapse problem by paralleling the original identity shortcut with more parameterized projections. The MSA module equipped with $T$ augmented shortcuts can be formulated as:
\begin{equation}
	\begin{aligned}
		\label{eq-augmsa}
		&{\rm AugMSA} (\boldsymbol{Z}_{l})=\MSA(\boldsymbol{Z}_l) + \boldsymbol{Z}_l  +  \sum_{i=1
		}^T \T_{{li}}(\boldsymbol{Z}_{l};\boldsymbol{\Theta}_{li}),\\
		&l \in [1,2,\cdots, L],
	\end{aligned}
\end{equation}
where $\T_{{li}}(\cdot)$ is the $i$-th augmented shortcut connection of the $l$-th layer and $\Theta_{li}$ denotes its parameters. In addition to the original shortcut, the augmented shortcuts provide more alternative paths to bypass the attention mechanism. Different from the identity projection that directly copies the input tokens to the corresponding outputs, the parameterized projection $\T_{{li}}(\cdot)$ transforms input features into another feature space. Projections $\T_{{li}}(\cdot)$ will apply different transformations to the input feature  as long as their weight matrices $\Theta_{li}$ are different, and thus paralleling more augmented shortcuts has the potential to enrich the feature space.

A simple formulation for  $\T_{li}(\cdot)$ is the sequence of a linear projection and an activation function  \ie, 
\begin{equation}
	\begin{aligned}
		\label{eq-aug}
		&\T_{li}(\boldsymbol{Z}_{l};\boldsymbol{\Theta}_{li}) = \sigma(\boldsymbol{Z}_l\boldsymbol{\Theta}_{li}),\\ 
		&l \in [1,\cdots, L],~i \in [1,2,\cdots, T],
	\end{aligned}
\end{equation} 
where $\Theta_{li}\in \R^{d\times d}$ is the weight matrix and $\sigma$ is the nonlinear activation function (\eg, GELU).
In Eq.~\ref{eq-aug}, $\T_{li}(\cdot)$ tackles each token independently and preserves their specificity, which is complementary to the MSA modules aggregating different tokens. Note that the identity mapping is a special case of Eq.~\ref{eq-aug}, \ie, $\sigma(x)=x$ and $\boldsymbol{\Theta}_{li}$ is the identity matrix.


This indicates that the upper bound of feature diversity $d_{\mathcal{M}_d}(\boldsymbol{Z}_l)$ decreases dramatically as the network depth increases without shortcut connections. We now analyze how the diversity $d_{\mathcal{M}_d}(\boldsymbol{Z}_l)$ changes \wrt the layer $l$ in the model stacked by the AugMSA modules. We have the following theorem.

\begin{theorem} 
	\label{th5-aug} 
	Given a model stacked by the AugMSA modules,  the diversity $d_{\mathcal{M}_d}(\boldsymbol{Z}_l)$ of feature in the $l$-th layer can be bounded by that of input data $\boldsymbol{Z}_0$, \ie,
	\begin{equation*}
		\small
		\begin{aligned}
			&d_{\mathcal{M}_d}({\rm AugMSA}(\boldsymbol{Z}_{l}))
			\leq (\sqrt{\lambda H} s\upsilon_1 + 1+ \sum_{i=1}^T L\Vert \boldsymbol{\Theta}_{li}\Vert_2)d_{\mathcal{M}_d}(\boldsymbol{Z}_l), \\
			&d_{\mathcal{M}_d}(\boldsymbol{Z}_{l})
			\leq (\sqrt{\lambda H} s\upsilon_1 + 1+ \sum_{i=1}^T L\Vert \boldsymbol{\Theta}_{li}\Vert_2)^ld_{\mathcal{M}_d}(\boldsymbol{Z}_0), 
		\end{aligned}
	\end{equation*}
	where $H$ is number of heads, $s>0$ is the largest element of all singular values of all $\boldsymbol{W}_l$, and $\Vert\cdot\Vert_2$ is the $\ell_2$ norm of the matrix.
\end{theorem} 


Since $\alpha_i = (\sqrt{\lambda H} s\upsilon_1 + 1+ \sum_{i=1}^T L\Vert  \boldsymbol{\Theta}_{li}\Vert_2) > 1$, this allows us to prevent feature collapse.


Compared with Theorem~\ref{th2-plain}, the augmented shortcuts introduce an extra term $(\sqrt{\lambda H} s\upsilon_1 + 1+ \sum_{i=1}^T L\Vert \boldsymbol{\Theta}_{li}\Vert_2)^l$, which increases exponentially. This tends to suppress the diversity decay incurred by the attention mechanism. The term $\alpha_i$ ($0\le i\le l$) is determined by the norms of weight matrices $\boldsymbol{\Theta}_{li}$ of the augmented shortcuts in the $i$-th layer, and the bound of diversity $d_{\mathcal{M}_d}(\boldsymbol{Z}_{l})$ in the $l$-th layer can be affected by all the augmented shortcuts in the previous layers. For the ShortcutMSA module with only an identity shortcut, we have $\alpha_i=\sqrt{\lambda} Hs\upsilon_1 + 1$. Adding more augmented shortcuts can increase the magnitude of $\alpha_i$, which further improves the bound. 

As discussed above, paralleling multiple augmented shortcuts with the MSA and MLP modules in a vision Transformer can improve the feature diversity to achieve higher performance. However, directly implementing $\T_{li}(\cdot)$~(Eq.~\ref{eq-aug}) involves a lot of matrix multiplications that are computationally expensive. For example, given feature $\boldsymbol{Z}_l \in \R^{n\times d}$ and weight matrix $\boldsymbol{\Theta}_{li} \in \R^{d \times d}$, the matrix multiplication $\boldsymbol{Z}_l\boldsymbol{\Theta}_{li}$ consumes $nd^2$ FLOPs, where $d$ is usually large in vision Transformers (\eg, 4096 in LLaMA-7B). In~\cite{tang2021augmented}, the augmented shortcut is implemented with a block-circulant matrix, which realizes fast inference with fast Fourier transformation (FFT). Even though it achieves high theoretical acceleration, we empirically find that its practical speed depends on the hardware optimization. Considering that LLM is a universal model, we propose to implement the augmented shortcut with a simpler bottleneck module. The module is constructed by stacking two FC layers with a nonlinear activation function (\eg, GeLU). The first FC layer reduces the $d$-dimension feature into a low-dimension space by a reduction ratio $r$ and the second FC layer restores the original feature dimension. Then the computational cost is reduced to $2nd^2/r$ FLOPs . A larger $r$ implies a further decrease in the computational cost. For example, when the reduction ratio $r$ is set to 32, the computational cost can be reduced by $16\times$ compared to the original augmented shortcut (Eq.~\ref{eq-aug}). 

Obviously, shortcuts use a large weight identity branch, such as ${\rm AugMSA} (\boldsymbol{Z}_l)= \MSA(\boldsymbol{Z}_l) + \alpha \boldsymbol{Z}_l(\sum_{i=1}^T \boldsymbol{\Theta}_{li})$ (where $\alpha > 0$), to prevent feature collapse, but this reduces network performance. We theoretically analyze this because feature diversity $d_{\mathcal{M}_d}(\boldsymbol{Z}_l)$ is adding excessive noise. The effect of noise on feature diversity is usually false positive. For example, if $\boldsymbol{H}=\boldsymbol{0}$, then $d_{\mathcal{M}_d}(\boldsymbol{H})=0$. However, when the input matrix introduces a zero-average noise $\boldsymbol{\epsilon}$, then $d_{\mathcal{M}_d}(\boldsymbol{H+\epsilon})=\Vert \boldsymbol{\epsilon} \Vert_F>d_{\mathcal{M}_d}(\boldsymbol{H})=0$. This requires us to improve the diversity features and minimize the impact of noise diversity on the network. This means reducing the value of $|d_{\mathcal{M}_d}({\rm AugMSA}(\boldsymbol{Z}_{l}+\boldsymbol{\epsilon})) - d_{\mathcal{M}_d}({\rm AugMSA}(\boldsymbol{Z}_{l}))|$ while ensuring that $d_{\mathcal{M}_d}(\boldsymbol{Z}_{l})$ is not attenuated.

The following describes the definition of feature diversity of noise.
We consider the effect of noise on matrix projection $d_{\mathcal{M}_d}(\boldsymbol{H})$.

\begin{equation}
	\begin{aligned}
		&|d_{\mathcal{M}_d}(\boldsymbol{H+\epsilon}) - d_{\mathcal{M}_d}(\boldsymbol{H})| 
		\leq\Vert \boldsymbol{\epsilon} -\boldsymbol{1}{(\boldsymbol{x}^{\boldsymbol{H+\epsilon}}_{\min}-\boldsymbol{x}^{\boldsymbol{H}}_{\min})}^\top \Vert_F \\
		\leq &d_{\mathcal{M}_d}(\boldsymbol{\epsilon})= \Vert(\boldsymbol{I}-\boldsymbol{ee}^\top )\boldsymbol{\epsilon}\Vert_F 
		= \Vert \boldsymbol{\epsilon} -\boldsymbol{1}{\boldsymbol{x}^{\boldsymbol{\epsilon}}_{\min}}^\top\Vert_F 
		\leq  \Vert \boldsymbol{\epsilon} \Vert_F.
	\end{aligned}
\end{equation}

For zero average noise, the following equation holds $\boldsymbol{ee}^\top \boldsymbol{\epsilon} = \boldsymbol{0} \in \mathbb{R}^{N\times d}$, so that the above inequality is equal. Since $|d_{\mathcal{M}_d}(f(\boldsymbol{H+\epsilon})) - d_{\mathcal{M}_d}(f(\boldsymbol{H}))| \leq d_{\mathcal{M}_d}(f(\boldsymbol{H+\epsilon})-f(\boldsymbol{H}))$. We define $d_{\mathcal{M}_d}(f(\boldsymbol{H+\epsilon})-f(\boldsymbol{H}))$ to represent the diversity effect of noise $\boldsymbol{\epsilon}$ on the $f$ function whose input is $\boldsymbol{H}$. The smaller the value, the higher the robustness of the function $f$. For the sake of simplicity and considering typical scenarios, the following discussion assumes that the input noise $\boldsymbol{\epsilon}$ is zero average noise.

\begin{lemma} 
	We consider the impact of noise on the MSA module, when $H=1$. For a slight perturbation of the input $\boldsymbol{\epsilon}$, the self-attention matrix also produces a perturbation $\boldsymbol{A_{\epsilon}}=\boldsymbol{A+\delta}$, \ie,
	\begin{align*}
		&d_{\mathcal{M}_d}({\rm MSA}(\boldsymbol{Z}_{l}+\boldsymbol{\epsilon}) -{\rm MSA}(\boldsymbol{Z}_{l}))\\
		&\leq \sqrt{\lambda_{\boldsymbol{A+\delta}}} s\upsilon_1 \Vert  \boldsymbol{\epsilon} \Vert_F + \sqrt{\lambda_{\boldsymbol{\delta}}}s\upsilon_1 d_{\mathcal{M}_d}(\boldsymbol{Z_{l}}),
	\end{align*}
	where $\lambda_{\boldsymbol{A+\delta}}$ is the largest eigenvalue of $\boldsymbol{{A_{\epsilon}}^\top}(\boldsymbol{I}-\boldsymbol{ee}^\top)\boldsymbol{A_{\epsilon}}$ and  $\lambda_{\boldsymbol{\delta}}$ is the largest eigenvalue of $\boldsymbol{{\delta}^\top}(\boldsymbol{I}-\boldsymbol{ee}^\top)\boldsymbol{\delta}$, usually $\lambda_{\boldsymbol{A+\delta}}<1$ and $\lambda_{\boldsymbol{\delta}}<1$.
	
	For the H heads MSA module, we can get the following formula:
	\begin{align*}
		&d_{\mathcal{M}_d}({\rm MSA}(\boldsymbol{Z_{l}+\epsilon}) -{\rm MSA}(\boldsymbol{Z_{l}}))\\
		&\leq \sqrt{\lambda_{\boldsymbol{A+\delta}}H} s\upsilon_1 \Vert  \boldsymbol{\epsilon} \Vert_F + \sqrt{\lambda_{\boldsymbol{\delta}}H}s\upsilon_1 d_{\mathcal{M}_d}(\boldsymbol{Z}_{l}),\\
	\end{align*}
\end{lemma} 

	\begin{lemma} 
		\label{lm4}
		We consider the noise diversity of linear parallel branch:
		\begin{align*}
			&d_{\mathcal{M}_d}(L (\boldsymbol{Z}_l+\boldsymbol{\epsilon})\boldsymbol{\Theta}_{li}- L\boldsymbol{Z}_l\boldsymbol{\Theta}_{li}) \leq L  \Vert \boldsymbol{\Theta}_{li} \Vert_2\Vert\boldsymbol{\epsilon} \Vert_F,
		\end{align*}
		
	\end{lemma}

\begin{theorem} 
	\label{th5} 
	If and only if $\boldsymbol{ee}^\top (\sigma(\boldsymbol{Z}_{l}+\boldsymbol{\epsilon}) -\sigma(\boldsymbol{Z}_{l})) = \boldsymbol{0}$, the following inequality is equal:
	\begin{align*}
		&d_{\mathcal{M}_d}(\T_{{li}}(\boldsymbol{Z}_{l}+\boldsymbol{\epsilon};\boldsymbol{\Theta}_{li})-\T_{{li}}(\boldsymbol{Z}_{l};\boldsymbol{\Theta}_{li}))\\
		&\leq L \Vert \boldsymbol{\Theta}_{li} \Vert_2 \Vert \boldsymbol{\epsilon} \Vert_F.
	\end{align*}
	For the nonlinear activation function, $\sigma(\boldsymbol{Z}_{l}+\boldsymbol{\epsilon}) -\sigma(\boldsymbol{Z}_{l})$ is no longer guaranteed to be zero-average. Therefore, the noise diversity of the nonlinear branch is weaker than that of the linear branch:
	\begin{align*}
		d_{\mathcal{M}_d}(\T_{{li}}(\boldsymbol{Z}_{l}+\boldsymbol{\epsilon};\boldsymbol{\Theta}_{li})-\T_{{li}}(\boldsymbol{Z}_{l};\boldsymbol{\Theta}_{li})) < L\Vert \boldsymbol{\Theta}_{li} \Vert_2\Vert \boldsymbol{\epsilon} \Vert_F.
	\end{align*}
	
\end{theorem} 

\begin{theorem} 
	\label{th6}
	Given a model stacked by the AugMSA modules, the noise diversity of feature in the $l$-th layer can be bounded by the following formula, \ie,
	\begin{equation*}
		\scriptsize
		\begin{aligned}
			&d_{\mathcal{M}_d}({\rm AugMSA}(\boldsymbol{Z}_{l}+\boldsymbol{\epsilon}) - {\rm AugMSA}(\boldsymbol{Z}_{l}))\\
			&< (1+\sqrt{\lambda_{\boldsymbol{A+\delta}}H} s\upsilon_1  + L \sum_{i=1}^T \Vert \boldsymbol{\Theta}_{li} \Vert_2)\Vert  \boldsymbol{\epsilon} \Vert_F + \sqrt{\lambda_{\boldsymbol{\delta}}H}s\upsilon_1 d_{\mathcal{M}_d}(\boldsymbol{Z}_{l})\\
		\end{aligned}
	\end{equation*}
\end{theorem} 

This indicates that using a large number of nonlinear shortcuts instead of $ L\boldsymbol{Z}_l(\sum_{i=1}^T \boldsymbol{\Theta}_{li}) $ prevents feature collapse, reduces the impact of input noise on feature diversity, and enhances the robustness of the network. In addition, it also enhances the nonlinear expression ability.

\subsection{Series Informed Activation Function}

A neural network $N_d$ composed of d hidden layers can be regarded as a composite of d functions $f_i$: $N_d = f_1 \circ f_2 \circ \cdots \circ f_d$.
In particular, each hidden layer function $f_i$ can be written as the composite of a function $g_i$ and an activation function $\sigma_i$: $f_i = \sigma_i \circ g_i$. Actually, the learning procedure of $f_i$ amounts to an optimization problem over the layer hypothesis space $H_i$. 

Usually, $\phi_i$ is taken as a non-learnable function; therefore, in the most common scenario, $H_i={\sigma_i}\times H_{g_i}$. $g_i$ is parameterized and learnable, and belongs to a hypothesis space $H_{g_i}$. This clearly limits the learnable space. 

In this section, we introduce a technique to define learnable activation functions that could be plugged into all the hidden layers of MLP. We define the hypothesis space $H_{\phi_i}$, based on the following idea: (i) select a finite set of activation functions $\Sigma := \{\sigma_1, \cdots , \sigma_N \}$, whose elements will be used as base elements; (ii) define the learnable activation function $\phi_i$ as a linear combination of the elements of $\Sigma$; (iii) identify a suitable hypothesis space $H_{\phi_i}$; (iv) optimize the whole network, where the hypothesis space of each hidden layer is $H_i = H_{\phi_i} \times H_{g_i}$. In this way, we expand the learnable space of each hidden layer, enhancing the model's nonlinear expression capability.  

Several different activation functions have been proposed for deep neural networks, including the most popular Rectified Linear Unit (ReLU) and its variants (PReLU~\cite{he2015delving}, GeLU~\cite{hendrycks2016gaussian} and Swish~\cite{ramachandran2017searching}). They focus on enhancing the performance of deep and complex networks using different activation functions. However, as theoretically proven in the preceding section, the limited power of Transformer architecture is mainly due to poor nonlinearity, which has not been fully investigated by the existing activation functions. 

There are two ways to improve the nonlinearity of a neural network: stacking the nonlinear activation layers or increasing the nonlinearity of each activation layer. The trend of existing networks is to choose the former, which results in high latency when there is excessive parallel computation ability.

One straightforward idea to improve the nonlinearity of the activation layer is stacking. The serial stacking of the activation function is the key idea of deep networks. However, stacking layers serially results in a large computation cost, which is not affordable for developing an efficient and effective LLM. Therefore, we choose concurrently stacking the activation function. Denote there are $n$ activation functions for input $x$ in a neural network as $\{\sigma_i(x)\}^n_{i=1}$, which can be the usual functions such ReLU and Tanh. The concurrent stacking of the activation functions can be formulated as:
\begin{equation}
	\sum_{i=1}^n  \sigma_i (a_i x+b_i),
	\label{equ1}
\end{equation}
where $n$ denotes the number of stacked activation functions and $a_i,b_i$ are the scale and bias (which are learned parameters) of each activation to prevent simple accumulation. 
The nonlinearity of the activation function can be largely enhanced by concurrent stacking. Equation~\ref{equ1} can be regarded as a series in mathematics, which is the operation of adding many quantities. 

Since the nonlinearity of the Transformer is mainly derived from the feed-forward network (FFN), we apply the series informed activation function on the FFN block. Given an input feature $x\in\mathbb{R}^{N \times D}$, where $N$ and $D$ are the number of tokens and its hidden dimension, the original FFN can be formulated as 
\begin{equation}
	\MLP (\boldsymbol{Z}'_{l}) = \sigma(\boldsymbol{Z}'_{l}\boldsymbol{W}'_{{l_1}})\boldsymbol{W}'_{{l_2}_i},~l \in [1,2,\cdots, L],  
\end{equation}
where $\boldsymbol{W}'_{{l_1}}$ and $\boldsymbol{W}'_{{l_2}}$ are two fully connected layers. Specifically, to further enrich the approximation ability of the series, we enable the series-based function to learn the global information by varying the inputs from their neighbors, which can be reformulated as:
\begin{equation}
	{\rm SIAF}-\MLP (\boldsymbol{Z}'_{l}) = (\sum_{i=1}^n \sigma_i(\boldsymbol{Z}'_{l}\boldsymbol{W}'_{{l_1}_i}))\boldsymbol{W}'_{{l_2}_i},~l \in [1,2,\cdots, L].  
\end{equation} 
It is easy to see that when $n=1$, the series-based activation function $\sigma_s(x)$ degenerates to the plain activation function $\sigma(x)$, which means that the proposed method can be regarded as a general extension of existing activation functions. 

\begin{theorem} 
	\label{th-full} 
	Given a model stacked by the ${\rm SIAF}-\MLP$ modules,   the diversity $d_\mathcal{M}(\boldsymbol{Z}'_{l})$ of feature in the $l$-th layer can be bounded by that of input data $\boldsymbol{Z}'_{0}$, \ie
	\begin{align*}
		&d_{\mathcal{M}_d}({{\rm SIAF}-\MLP}(\boldsymbol{Z}'_{l})) \leq (\sum_{i=1}^n L_i)s\upsilon_2 d_{\mathcal{M}_d}(\boldsymbol{Z}'_{l}),\\
		&d_{\mathcal{M}_d}(\boldsymbol{Z}'_{l})
		\leq (s\upsilon_2\sum_{i=1}^n L_i)^ld_{\mathcal{M}_d}(\boldsymbol{Z}'_{0}), 
	\end{align*}
	where $L_i$ is the Lipschitz constant of the activation function $\sigma_i$. 
\end{theorem} 

As demonstrated in our proof, the Series Informed Activation Function (SIAF) we propose markedly amplifies the nonlinear expressive capacity of the MLP module compared to the original architecture. This enhancement in nonlinearity progressively intensifies with the increase of the parameter $n$.

\subsection{Combination}

Finally, we offer the upper bounds for the combination of multi-layer AugMSA module and SIAF-MLP module to decay into subspace $\mathcal{M}_d$.
We can obtain the upper bounds for the combination of multi-layer MSA module and MLP module to decay into subspace $\mathcal{M}_d$ in vanilla Transformer architecture.

\begin{theorem} 
	\label{th3-full} 
	Provide a network consisting of p-layer MSA module and q-layer MLP module,  the diversity $d_{\mathcal{M}_d}(\boldsymbol{Z}_{p+q})$ of feature in the $l$-th layer can be bounded by that of input data $\boldsymbol{Z}_0$, \ie, 
	\begin{align}
		&d_{\mathcal{M}_d}(\boldsymbol{Z}_{p+q})\leq (\sqrt{\lambda H} s\upsilon_1)^{p}(Ls\upsilon_2)^{q}d_{\mathcal{M}_d}(\boldsymbol{Z}_0). 
	\end{align}
\end{theorem}

It is evident that the original Transformer architecture possesses a relatively limited upper bound in terms of nonlinear expressive capability. Building upon this observation, we now proceed to analyze the enhanced expressive power of the Transformer  when augmented with our proposed architectural modifications.

\begin{theorem} 
	\label{th9-full} 
	Provide a network consisting of p-layer AugMSA module and q-layer SIAF-MLP module,  the diversity $d_{\mathcal{M}_d}(\boldsymbol{Z}_{p+q})$ of feature in the $l$-th layer can be bounded by that of input data $\boldsymbol{Z}_0$, \ie, 

		\begin{align*}
			&d_{\mathcal{M}_d}(\boldsymbol{Z}_l) \\
			&\leq (\sqrt{\lambda H} s\upsilon_1 + 1+ \sum_{i=1}^T L\Vert \boldsymbol{\Theta}_{li}\Vert_2)^p(s\upsilon_2\sum_{i=1}^n L_i)^{q}d_{\mathcal{M}_d}(\boldsymbol{Z}'_0). 
		\end{align*}
\end{theorem} 

The preceding theorem substantiates that our proposed augmented shortcut module, when used in conjunction with the series informed activation function module, enhances the model's nonlinear expressive capabilities and diversity far beyond what is achievable by using either module independently. Consequently, our Transformer architecture amalgamates these two modules to form our PanGu-$\pi$ architecture, resulting in a synergistic improvement in nonlinearity and diversity.

\section{Experiments on General Field}

In this section, we compare the existing open-source 7B and 1B models. Furthermore, we conduct ablation studies of the proposed architecture. 

\textbf{Training data} The pre-training data is gathered from diverse sources from the Internet, covering English and Chinese corpus in an equal $1:1$ ratio. The tokenizer is built by byte-pair encoding (BPE)~\cite{shibata1999byte} from SentencePiece~\cite{kudo2018sentencepiece} upon our data. The final vocabulary size is about 0.1 million. After tokenization, the entire training dataset contains about 1.6 trillion tokens. 

\textbf{Training details} Our models are trained using the AdamW optimizer~\cite{loshchilov2017decoupled} with $\beta_1 = 0.9, \beta_2 = 0.95$ for 1 epoch utilizing the cosine learning rate decay~\cite{loshchilov2016sgdr} with an initial learning rate $3\times 10^{-4}$. The total batch size for the training process is approximately 4M, and it includes a warm-up phase spanning 4000 steps.

\textbf{Model details} For fair comparison, we adopt the pre-normalization~\cite{zhang2019root}, SwiGLU activation~\cite{shazeer2020glu} and rotary embeddings~\cite{su2023roformer} following the LLaMA architecture~\cite{touvron2023llama}. We then apply our series activation function and augmented shortcuts to build our models. The details of the models can be found in Table~\ref{tab:model}. We reduce the number of layers to make the number of parameters similar to the LLaMA model for fair comparison because the proposed modules introduce extra parameters.

\textbf{Training Devices} We use the Huawei Ascend 910A card to train and evaluate the proposed architecture. 
The HUAWEI Ascend 910A is a high-efficiency, flexible, and programmable artificial intelligence processor. For half-precision floating-point (FP16) operations, the Ascend 910 delivers 256 TeraFLOPS. For integer precision calculations (INT8), it delivers 512 TeraOPS. Despite its unparalleled performance, Ascend 910's maximum power consumption is only 310W, which is significantly lower than its planned specifications of 350W. Developed using Huawei's proprietary Da Vinci architecture, it integrates a rich array of computing units, enhancing the completeness and efficiency of AI computations, thereby extending its applicability. It significantly improves the performance of the entire AI system and effectively reduces deployment costs.

\begin{table}[h]
	\centering
	\caption{Model details of PanGu-$\pi$.}
	\label{tab:model}
	\begin{tabular}{l|c|c|c}
		\toprule[1.5pt]
		Models & dimension & $n$ heads & $n$ layers  \\
		\midrule
		PanGu-$\pi$-1B &   2048 & 16 & 12 \\
		PanGu-$\pi$-7B &  4096 & 32 & 29  \\
		\bottomrule[1pt]
	\end{tabular}
\end{table}

\textbf{Benchmarks} We use the {\tt{OpenCompass}} platform~\cite{2023opencompass} to evaluate on an extensive suite of downstream tasks. We selected 11 classical benchmarks from four different domains to conduct a comprehensive comparison.  C-Eval~\cite{huang2023c} is a comprehensive Chinese evaluation benchmark to evaluate the knowledge and reasoning abilities of LLMs, which includes multiple-choice questions from 52 diverse disciplines across different difficulty levels. CMMLU~\cite{li2023cmmlu} is also a comprehensive Chinese evaluation benchmark, which covers 67 topics including science, engineering, and humanities. MMLU~\cite{hendrycks2020measuring} proposes an English evaluation benchmark for measuring LLM's multitask accuracy by covering 57 tasks including mathematics, history, computer science, and law. AGI-Eval~\cite{zhong2023agieval} is a human-centric benchmark specifically designed to evaluate the general abilities of foundation models in tasks pertinent to human cognition and problem-solving. BoolQ~\cite{clark2019boolq} is a reading comprehension dataset to evaluate the difficult entailment-like inference ability of LLMs. AX-b~\cite{wang2020superglue} is a broad-coverage diagnostic task and PIQA~\cite{bisk2019piqa} is a physical interaction question-answering task. CSL~\cite{li2022csl} offers a Chinese Scientific Literature dataset to evaluate the performance of models across scientific domain tasks. EPRSTM~\cite{xu2021fewclue} is a binary sentiment analysis dataset based on product reviews on an e-commerce platform. ~\cite{narayan2018dont} is a single-document summarization task, and LCSTS~\cite{hu2016lcsts} is a large corpus of Chinese short-text summarization datasets.

\subsection{Ablation Studies}

To better understand the proposed Architecture, we conduct extensive experiments to investigate the impact of each component. All the ablation experiments are conducted based on the 1B model size. 

\begin{table}[h]
	
	\centering
	\caption{Ablation Study about series informed activation function.}
	\begin{tabular}{c|c|c}
		\toprule
		Number of $n$        & C-Eval & Inference Speed (ms) \\
		\midrule
		1           & 35.0    & 7.56            \\
		2         & 36.9    &    7.76      \\
		3        & 37.0    & 8.02         \\
		4           & 37.0    & 8.35            \\
		\bottomrule
	\end{tabular}

	\label{tab:series}
	
\end{table}

\textbf{Influence of series informed activation function.} In the above section, we propose the SIAF to enhance the performance and enable global information exchange in feature maps. Table~\ref{tab:series} shows the performance of the proposed SIAF using different numbers of $n$. When $n=1$, the activation function degenerates into the plain activation function. We find that when $n=2$, the performance and the inference speed strike an optimal balance. Therefore, we choose $n=2$ for the following experiments.

\begin{table}[h]
	\centering
	\caption{Influence of the width factor in augmented shortcuts.}
	\begin{tabular}{c|c|c}
		\toprule
		Width factor of bottlenecks        & C-Eval & Inference Speed (ms) \\
		\midrule
		1 & 36.5    & 8.66         \\
		1/4          & 36.7    & 8.06            \\
		1/16        & 36.5    &  7.96      \\
		1/32       & 36.3    & 7.66         \\
		1/64       & 35.5    & 7.62         \\
		0      & 35.0    & 7.56         \\
		\bottomrule
	\end{tabular}
	\label{tab:aug}
\end{table}

\textbf{Influence of augmented shortcuts.} As mentioned above, the augment module can greatly improve the performance. As shown in Table~\ref{tab:aug}, we trade off accuracy and speed by controlling the width of the bottleneck middle layer. By transforming the reduction rate, it is apparent that an increase in reduction rate results in a decrease in calculation speed and accuracy. After careful consideration, we determined a reduction rate of 32 to achieve the optimal balance between speed and accuracy.

\begin{table}[h]
	\centering
	\caption{Ablation Study of each component.}
	\begin{tabular}{c|c|c}
		\toprule
		Method        & C-Eval & Inference Speed (ms) \\
		\midrule
		Vanilla Transformer & 35.0    & 7.56         \\
		WideNet~\cite{xue2022go} & 36.5    & 8.06         \\
		SIAF          & 36.9    & 7.76            \\
		AS         & 36.3    &    7.66      \\
		SIAF+AS       & 37.9    & 7.86         \\
		\bottomrule
	\end{tabular}
	\label{tab:abl}
\end{table}

\textbf{Architecture} We finally ablate the effectiveness of each component of the proposed method and report the language modeling results in Table~\ref{tab:abl}. We ablate the series informed activation function (SIAF) and augmented shortcuts (AS) as described earlier. Furthermore, we compared the proposed method with WideNet~\cite{xue2022go}, which is introduced to increase the nonlinearity of Transformer architectures. It can be seen through experiments that each component of the proposed method is effective for improving the performance of the Transformer architecture, which surpasses that of  WideNet~\cite{xue2022go}.\

\subsection{Feature Analysis and Visualization}
\begin{figure}[!h]
	\centering
	\includegraphics[width=0.7\linewidth]{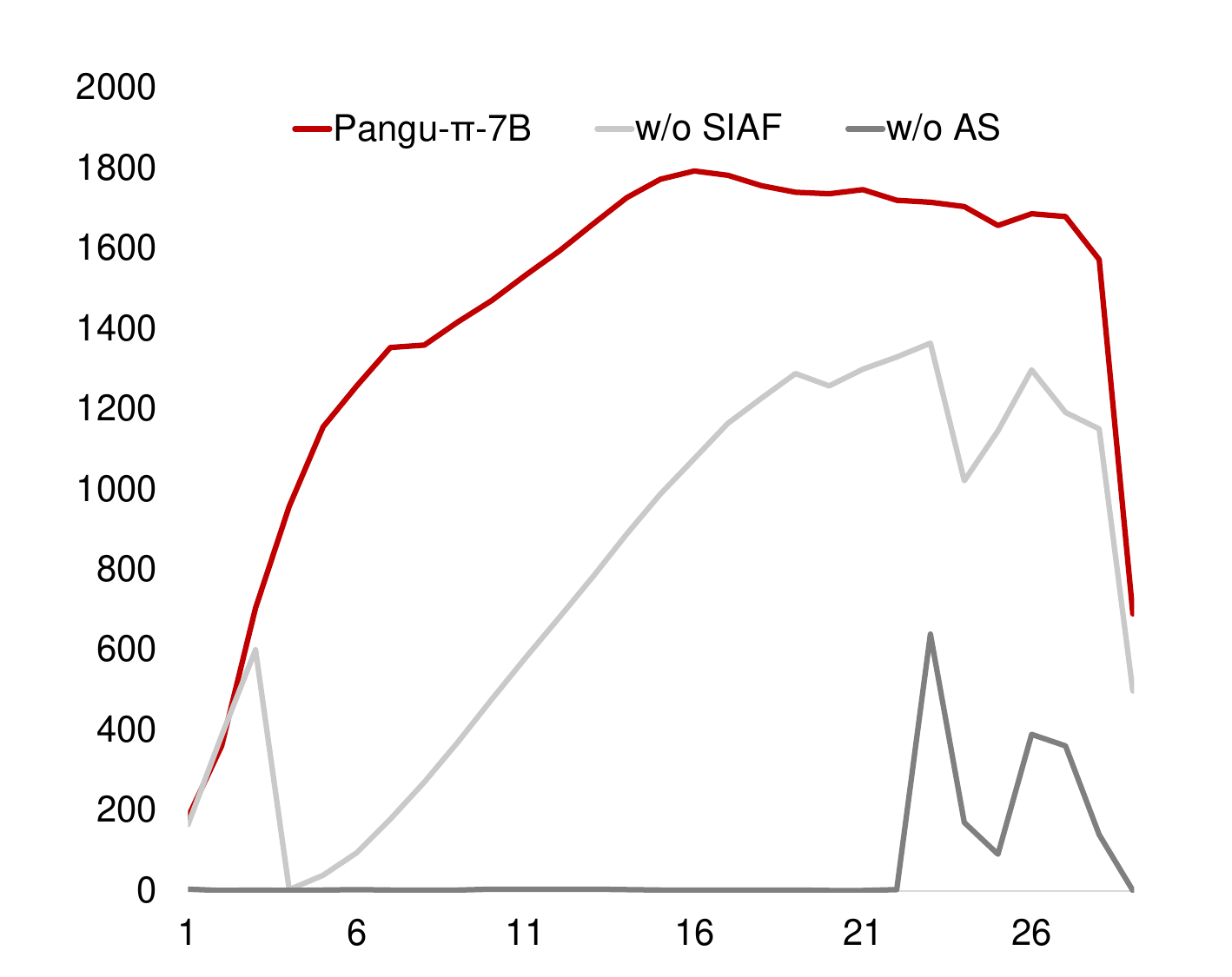}
	\caption{The effective dimension $d(0.8)$ across layers of different model architectures. A larger number of effective dimensions means more principal components are needed to account for 80\% of variance, indicating more diversity in feature channels.}
	\label{eff-dim}
\end{figure}

We also analyze the latent representation across different layers to demonstrate further the superiority of nonlinearity compensation introduced by the PanGu-$\pi$ architecture. We are interested in the channel diversity of each architectural choice. Following the analysis method from~\cite{Cai2021IsotropyIT}, we characterize the effective dimensions of different decoder architectures. In particular, the effective dimension $d(\epsilon)$ is defined as the minimum principal component numbers that occupy the explained variance ratio of $\epsilon$ in a principal component analysis (PCA). In principle, a more powerful representation of individual tokens would result in a larger effective dimension. In comparison, a smaller effective dimension means the variance in token representation occurs mainly in smaller dimensions. As shown in Fig.~\ref{eff-dim}, here we report each layer's effective dimension $d(0.8)$. Removing all augmented shortcuts limits the effective dimension to the greatest extent, and removing the series informed activation function significantly reduced the effective dimension consistently on each Transformer layer, indicating the significant role of these components in channel-wise feature diversity~\cite{Ethayarajh2019HowCA}.

\begin{figure}[!h]		
	\centering	
	\includegraphics[width=1.0\linewidth]{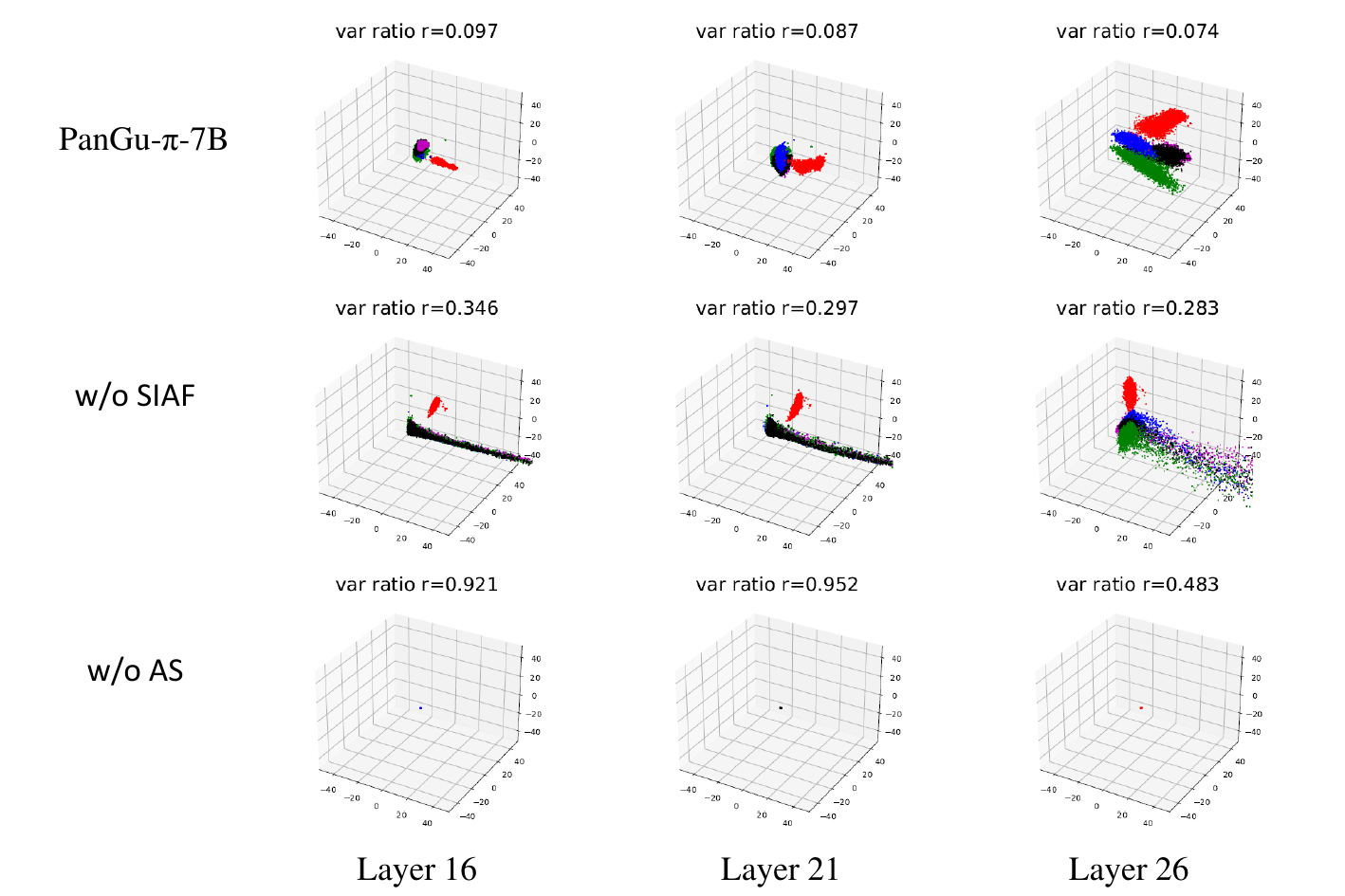}	
	
	\caption{Visualization of hidden states from each layer. The most frequent five tokens are highlighted using different colors for visualization. The total variance accounted for by the first three principal components is labeled for each layer on the top. Note that the beginning tokens are removed from the analysis because they are considered outliers.}	
	\label{vis-3d}	
\end{figure}

Furthermore, to offer a finer-grained characterization of linguistic features from different architectures, we also visualize the representation of tokens with different semantics, using the test set of Penn Tree Bank (PTB)~\cite{2002Building} as a general domain corpus. In particular, we adopt a layer-wise visualization method to illustrate the concentration and diversity of features for each token and how these characteristics change along the Transformer layers Fig.~\ref{vis-3d}. To assist the visualization, the top five frequent tokens are highlighted with different colors. PCA is used to reduce all feature maps to a 3D space, preventing nonlinear reduction as it may cause disruption. Additionally, the total variance accounted for by the first three principal components is labeled for each layer. From Fig.~\ref{vis-3d}, we can see that PanGu-$\pi$ architecture possesses the most diverse and isotropic feature space~\cite{2021SimCSE}, with each token's feature expanding to a high-dimensional cluster when moving to deeper layers. In comparison, removing the series informed activation function or the augmented shortcuts limit the feature on low-dimensional manifolds (i.e., aggregating along one axis in the middle panel), and the differentiation among tokens is also blurred, indicating less discriminative power in language modeling.

\begin{table*}[htbp]
	\centering
	\caption{Comparison with SOTA open-source 7B models. The best model is listed in bold.}
	\label{tab:7b}
	\vspace{-4pt}
	\begin{tabular}{ccccccc}
		\toprule[1.5pt]
		& & \multicolumn{5}{c}{\textbf{Models}} \\
		\cmidrule{3-7}  \multicolumn{1}{l}{\textbf{Task}} & \textbf{Dataset} & LLaMA2-7B & Baichuan2-7B & InternLM-7B & Qwen-7B & PanGu-$\pi$-7B \\
		\midrule
		\multicolumn{1}{l}{\multirow{4}{*}{Examination}} & C-Eval & 32.50  & 56.90  & 53.21  & \textbf{63.40} & 59.86  \\
		& CMMLU & 31.80  & 57.02  & 51.86  & \textbf{62.50} & 59.92  \\
		& MMLU  & 46.80  & 54.72  & 51.39  & 59.70  & \textbf{61.91} \\
		& AGI-Eval & 21.80  & 34.77  & 37.77  & 45.30  & \textbf{54.16} \\
		\midrule
		\multicolumn{1}{l}{Knowledge} & BoolQ & 74.90  & 63.21  & 64.10  & \textbf{76.10} & 64.77  \\
		\midrule
		\multicolumn{1}{l}{\multirow{2}{*}{Reasoning}} & AX-b  & 53.50  & 51.72  & 42.57  & 57.00  & \textbf{57.52} \\
		& PIQA  & \textbf{78.30} & 76.22  & 78.02  & 77.90  & 77.15  \\
		\midrule
		\multicolumn{1}{l}{\multirow{4}{*}{Understanding}} & CSL   & 55.60  & \textbf{66.25} & 65.62  & 56.20  & 63.75  \\
		& EPRSTMT & 46.20  & 69.38  & 88.12  & 88.80  & \textbf{90.62} \\
		& XSum  & 19.70  & \textbf{20.89} & 8.12  & 1.30  & 19.58  \\
		& LCSTS & 9.10  & 15.57  & \textbf{18.19} & 12.00  & 16.62  \\
		\midrule
		\multicolumn{2}{c}{Average} & 42.75  & 51.19  & 50.82  & 54.56  & \textbf{56.90}  \\
		\midrule
		\multicolumn{2}{c}{Latency on 910A (ms)} & \multicolumn{4}{c}{47.60}    & \textbf{43.00}  \\
		\bottomrule[1pt]
	\end{tabular}%
\end{table*}%

\begin{figure}
	\centering
	\begin{tabular}{cc}
		\includegraphics[width=0.48\linewidth]{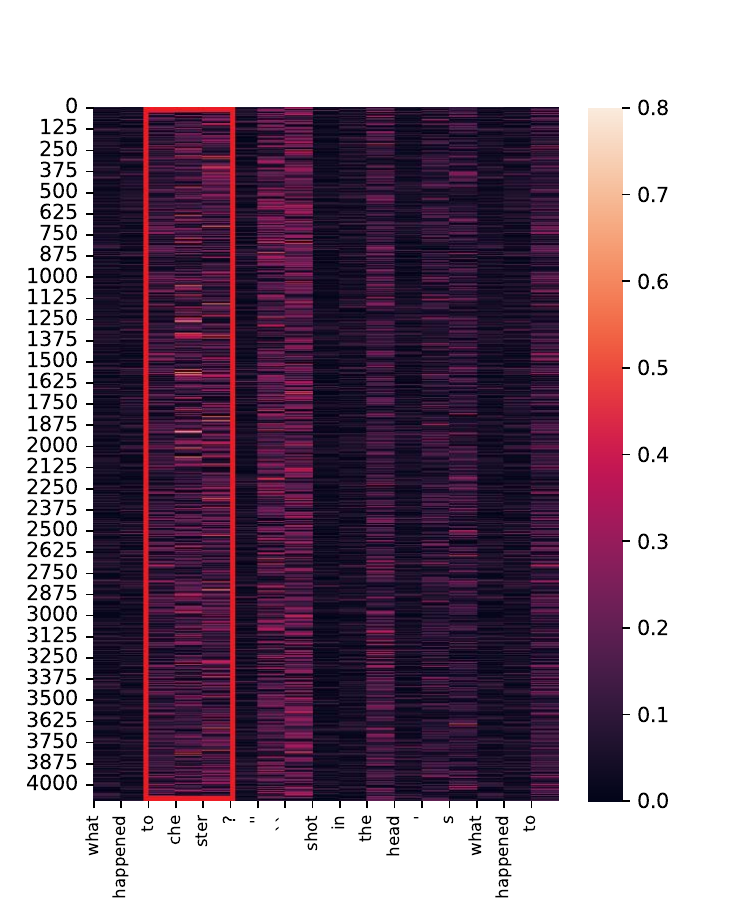} &
		\includegraphics[width=0.48\linewidth]{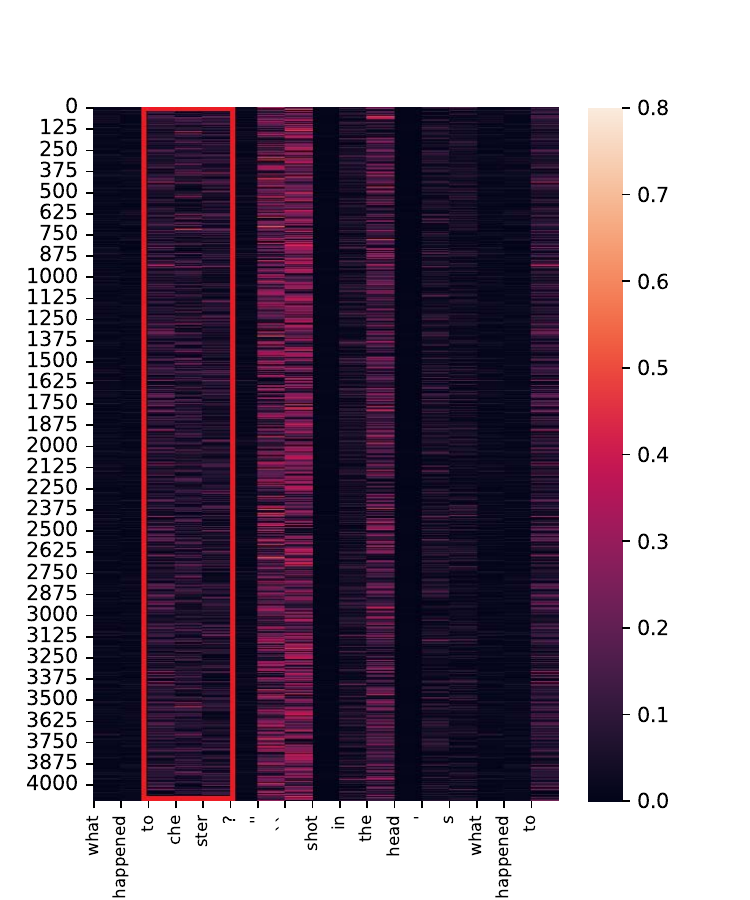} \\
		(a)  PanGu-$\pi$  & (b) Vanilla Transformer  \\
	\end{tabular}
	
	\caption{Saliency analysis for next word prediction task. Color brightness indicates the absoluate value of the corresponding gradients with respect to the prediction target. The gradient values are normalzied within each sample to ensure a fair comparison. Important messages are framed in red.}
	\label{case-sil}
\end{figure}

To verify the effectiveness of the language representation enhanced by PanGu-$\pi$ architecture, we conduct case analyses where the saliency of each token's feature dimension is calculated by deriving the absoluate value of the corresponding gradients with respect to the prediction target. 
As shown in Figure~\ref{case-sil}, the language model is required to echo the previous mentioned name ``chester" as the next word. The PanGu-$\pi$ model correctly identifies the key message ``chester" in the context reflected by higher gradient values for most of the channels (Figure~\ref{case-sil} (a)). In comparison, without the augmented shortcuts and series activation function, the model tend to leverage more information from the unmeaningful symbols after the key hints, leading to a wrong prediction that directly ends the sentence (Figure~\ref{case-sil} (b)).

\begin{table*}[htbp]
	\centering
	\caption{Comparison with SOTA open-source 1B models. The best model is listed in bold.}
	\label{tab:1b}
	\begin{tabular}{cccccc}
		\toprule
		&       & \multicolumn{4}{c}{\textbf{Models}} \\
		\cmidrule{3-6}    \multicolumn{1}{l}{\textbf{Task}} & \textbf{Dataset} & Sheared-LLaMA-1.3B & Chinese-LLaMA2-1.3B & TinyLLaMA-1.1B & PanGu-$\pi$-1B \\
		\midrule
		\multicolumn{1}{l}{\multirow{4}{*}{Examination}} & C-Eval & 24.28  & 28.70  & 27.85  & \textbf{36.85} \\
		& CMMLU & 25.10  & 24.78  & 24.64  & \textbf{35.90} \\
		& MMLU  & 25.77  & 24.55  & 25.75  & \textbf{35.96} \\
		& AGI-Eval & 18.01  & 19.40  & 18.54  & \textbf{30.77} \\
		\midrule
		\multicolumn{1}{l}{Knowledge} & BoolQ & \textbf{62.39} & 56.79  & 56.06  & 58.44  \\
		\midrule
		\multicolumn{1}{l}{\multirow{2}{*}{Reasoning}} & AX-b  & 43.57  & \textbf{47.46} & 45.47  & 43.48  \\
		& PIQA  & \textbf{72.91} & 56.91  & 70.62  & 61.92  \\
		\midrule
		\multicolumn{1}{l}{\multirow{4}{*}{Understanding}} & CSL   & 51.88  & 55.60  & 53.12  & \textbf{55.62} \\
		& EPRSTMT & 46.25  & \textbf{72.50} & 46.25  & 55.62  \\
		& XSum  & 16.44  & 8.90  & \textbf{20.15} & 15.92  \\
		& LCSTS & \textbf{15.37} & 13.16  & 13.97  & 14.61  \\
		\midrule
		\multicolumn{2}{c}{Average} & 36.54  & 37.16  & 36.58  & \textbf{40.46} \\
		\bottomrule
	\end{tabular}%
\end{table*}%

\subsection{Comparison with 7B Models}

To showcase the competitive edge of PanGu-$\pi$, we conducted an extensive analysis of the PanGu-$\pi$ model, comparing it against other state-of-the-art models with a similar size. The comprehensive results are detailed in Table~\ref{tab:7b}. We segmented the comparative datasets into four tasks: examination, knowledge, reasoning, and understanding, to fully assess the capabilities of our model. Notably, in the examination category, our model almost reached the state-of-the-art (SOTA) benchmarks, surpassing LLaMA2, Baichuan2, and InternLM, and closely matching the current top performer, Qwen. The trend of superior performance in reasoning and understanding categories can also be observed, which is attributed to the PanGu-$\pi$ model's advanced nonlinear capability. This allows it to fit more complex function spaces, giving it a competitive advantage in these challenging SOTA benchmarks.

In the knowledge dataset, our model still lags behind in performance when compared to the existing SOTA benchmarks. This could be due to the lower concentration of knowledge in our collected dataset and the presence of many unseen data points in the BoolQ dataset. Nevertheless, our model achieved excellent results, demonstrating strong generalization abilities.

Overall, our model exhibits consistently better average performance indices compared to the current state-of-the-art models. When measured against other open-sourced models of a 7B size, PanGu-$\pi$ achieves significantly higher average performance. In the future, we plan to train our model with better data, aiming to enhance its performance metrics in the knowledge domain.

Moreover, we evaluated the latency (milliseconds per token) of these models. As the compared models utilize a similar architecture to LLaMA, their latencies are comparable. Our findings indicate that PanGu-$\pi$ achieves a much faster inference speed compared to the LLaMA architecture, further establishing its superiority.

The PanGu-$\pi$ model can serve as a swift and effective foundational model, leveraging strategies such as  Supervised Fine-Tuning (SFT), AI Agent, or retrieval augmentation to become a competitive AI solution. It has the potential to harness the power of LLMs in edge devices like smartphones, offering a compelling choice for various applications.

\subsection{Comparison with 1B Models}
For comparison, we have meticulously selected three SOTA models of similar size, all based on the LLaMA architecture. These include Chinese-LLaMA2-1.3B~\cite{chinese-llama-alpaca}, TinyLlama-1.1B~\cite{tinyllama}, and Sheared-LLaMA-1.3B~\cite{xia2023sheared}. Notably, Sheared-LLaMA-1.3B was initially pruned from a larger LLaMA2-7B model and subsequently trained with a condensed dataset of 50B tokens. Our extensive experiments, as presented in Table~\ref{tab:1b}, demonstrate that PanGu-$\pi$-1B significantly surpasses existing LLMs of similar size, and in some cases, even larger sizes.

Similar to our findings with the 7B model, our 1B model demonstrates a remarkable superiority in the examination category over existing models, affirming the efficacy of our modifications even at the 1B scale. This outcome is a testament to our model’s enhanced nonlinear capabilities, enabling it to fit more complex function spaces. In the realms of reasoning and understanding, we observe similar trends, further validating our model's robust performance.

However, as with our larger model, the knowledge density in our collected dataset for training the 1B model was relatively low. Consequently, in datasets like BoolQ, which contain data points previously unseen by our model, we acknowledge a noticeable gap in performance when compared to current state-of-the-art benchmarks. This indicates room for improvement in our model's ability to handle unfamiliar data, a challenge we aim to address in future iterations.

Furthermore, when assessing latency, a critical factor in real-world applications, PanGu-$\pi$-1B shows a marked advantage. It records lower latency times of 13.8 ms on the 910A, compared to its deeper counterpart, LLaMA2-1B, which clocks 15.4 ms on the 910A, despite having a roughly equivalent number of parameters. This not only underscores the efficiency of PanGu-$\pi$-1B but also highlights its suitability for deployment in time-sensitive applications.


\section{YunShan: Domain Specialized Model for Finance and Legal fields}\label{Sec:YunShan}
In the current landscape of LLMs, there is a growing trend towards using domain-specific models for specialized applications, which differ from the capabilities of general-purpose large models. These domain-specific models are developed with a deeper level of expertise, and are particularly adept at addressing tasks in specific fields. For instance, in the finance industry, LLMs are used to provide in-depth analysis of financial documents and offer interactive information services for market insights and queries. Similarly, in the legal field, LLMs are used to facilitate expedited case research and information retrieval. Exemplary models such as Xuanyuan~\cite{zhang2023xuanyuan}, FinGPT~\cite{yang2023fingpt} and FinMA~\cite{xie2023pixiu} in financial services, and Lawyer LLaMA~\cite{huang2023lawyer} and LawGPT~\cite{LAWGPT-zh} for legal application. This progression not only exemplifies the robust development within the sphere of domain-specific LLMs but also signals the future role these technologies are poised to play in furthering innovation and growth across various domains.

\subsection{Datasets}
To enhance the model’s ability to solve domain-specific tasks, the model is continually pretrained and instruction finetuned on the domain datasets. We collect a variety of domain datasets both in the financial domain and the legal domain. When we construct the datasets, we consider the professionalism, diversity, data size and accessibility of the sources. We process and clean the collected datasets to construct high-quality domain corpora.

\noindent\textbf{Financial Domain} Our financial pretraining data consists of company announcements, financial news, articles, and examinations. Some of the data is from FinCorpus\footnote{https://huggingface.co/datasets/Duxiaoman-DI/FinCorpus\label{web}}, which is a high-quality Chinese financial dataset. The rest of the data is crawled through TuShare\footnote{https://tushare.pro/}, which is a Python package that provides free access to Chinese financial data from various sources. After cleaning, the total size of the data is 36.5B tokens.

\noindent\textbf{Legal Domain} Our legal pretraining data comprises legal regulations, legal case texts, academic papers, and legal examinations. We use Pile of Law~\cite{2022pileoflaw} and LeXFiles~\cite{2023-lexfiles} to construct some of the data, and selectively collect data from two legal-related public platforms\footnote{https://pkulaw.com/}\footnote{https://wenshu.court.gov.cn/}. After cleaning, the legal pretraining data contains 111.7B tokens.

\noindent\textbf{Instruction-following Data} For supervised fine-tuning, we collect 995k domain instruction-following data. The data consists of JEC-QA~\cite{zhong2019jec} and the instructions open sourced by ChatLaw~\cite{cui2023chatlaw}, Lawyer LLaMA~\cite{huang2023lawyer}, LawGPT~\cite{LAWGPT-zh}, and FinCorpus\textsuperscript{\ref {web}}, covering Judicial Examination examples, legal consultations, and financial examination examples.

\begin{table}
	\centering
	\caption{The vocab size and text compression rate of YunShan's tokenizer compared with other tokenizers of LLaMA2, InternLM, Baichuan2 and PanGu-$\pi$.}
	\begin{tabular}{c|c|c}
		\toprule
		Tokenizer        & Vocab Size & Compression Rate \\
		\midrule
		LLaMA2           & 32,000     & 1.479            \\
		InternLM         & 103,168    & 0.657            \\
		Baichuan2        & 125,696    & 0.605            \\
		\midrule
		\textbf{PanGu}            & 100,883    & 0.649            \\
		\textbf{YunShan} & 110,428    & 0.606            \\
		\bottomrule
	\end{tabular}
	
	\label{tab:vocabulary-compressionrate}
	
\end{table}

\subsection{Tokenizer}
Corpora from financial and legal fields often contain specialized words that are not included in the general domain vocabulary. During tokenization, these unknown words are deconstructed into sub-words or even UTF-8 bytes, which reduces model training efficiency. To alleviate the inefficiency, we expanded the original vocabulary size from 100883 to 110428 by adding a specialized financial-legal word list.
We used byte-pair encoding (BPE)~\cite{wang2020neural} from Tokenizer\footnote{https://github.com/huggingface/tokenizers} to train our tokenizer. The specialized vocabulary is trained from a subset of the financial and legal corpora and then merged into the original vocabulary. Table~\ref{tab:vocabulary-compressionrate} shows the compression rate we calculate on the JEC-QA dataset~\cite{zhong2020jec}, a legal domain QA dataset. Our new tokenizer has a much lower compression ratio than the original one and ties for the best with Baichuan 2 while having a smaller vocabulary.

\subsection{Training Process}

\noindent\textbf{Further Pretraining}
To transfer financial and legal knowledge to the PanGu-$\pi$ model, we further pre-train it with a specialized vertical corpus.
To alleviate catastrophic forgetting of formerly learned knowledge, we partially introduce the during the pre-training process. Specifically, we resample a portion of the previous high-quality data and mix it with the vertical domain data at a $1:1$ ratio.

\noindent\textbf{Instruction Tuning} During the instruction tuning phase, we utilize supervised fine-tuning samples of both general-purpose tasks and domain-specific tasks. We shuffle and combine them into one supervised dataset and execute the instruction tuning process in one stage. On the basis of the YunShan-base model, which has already acquired general and specialized knowledge during previous phase, this approach aims to simultaneously teach the model how to follow human instructions in different domains.

\subsection{Benchmarks}

The proposed YunShan model is evaluated across the two specialized law and finance domains. This approach aims to comprehensively assess YunShan model's capabilities in knowledge retention, comprehension, and application within domain-specific tasks, covering Chinese and English languages.

\begin{table*}
	\centering
	
	\footnotesize
	\caption{The experimental results of general-domain and domain specialized LLMs on financial FinancelQ~\cite{zhang2023xuanyuan} benchmark.}
	\begin{tabular}{cccccccccccc}
		\toprule
		Model&CPA & BQ & SPQ & CFPQ & IPQ& Economist &Tax Accountant & FQ & Financial Planner & Actuary & \textbf{AVG} \\ 
		\midrule
		LLaMA2-7B         &  28.38  & 26.53  &  26.11  &  24.54  & 26.44 & 25.77 &  28.41 &26.10 &25.00 &25.41  &26.27 \\
		InternLM-7B       & 31.81  & 43.57  &  34.69  &  40.02  & 36.49 & 42.50 &  43.19 &37.29 &32.95 &31.56 &37.41 \\
		Baichuan2-7B      &  34.25  & 49.04  &  40.14  &  42.32  & 41.52 & 52.69 &  41.11 &41.02 &30.68 &35.86 &40.86 \\
		Qwen-7B           &  30.36  & 37.38  &  35.97  &  37.61  & 33.19 & 43.65 &  38.11 &38.31 &27.27 &27.66 &34.95 \\
		\midrule
		FinGPT-7B         & 25.32  & 25.88  &  24.40  &  24.89  & 25.57 & 25.58 &  27.48 &20.68 &22.73 &29.51 &25.20 \\
		FinMA-7B          &  24.94  & 29.66  &  33.25  &  29.93  & 35.34 & 32.31 & 26.56 &26.78 &22.73 &25.00 &28.65 \\
		LawGPT-7B         &  26.47  & 24.92  &  23.98  &  24.77  & 25.14 & 24.62 &  27.02 &27.46 &20.45 &26.43 &25.13 \\
		HanFei-7B         &  27.00  & 34.08  &  33.93  &  34.52  & 37.36 & 37.88 &  32.79 &30.51 &22.73 &30.33 &32.11 \\
		YunShan (Ours) &  \textbf{52.78}  & \textbf{63.42}  & \textbf{ 52.89}  &  \textbf{57.57}  & \textbf{52.30} & \textbf{72.50 }&  \textbf{52.66} &\textbf{55.93} &\textbf{34.09} &\textbf{51.43} & \textbf{54.56} \\
		\bottomrule
	\end{tabular}
	
	\label{tab:experiment_finance1}
\end{table*}
\subsubsection{Financial Domain}

For the financial domain, we use two distinct datasets: FinanceIQ~\cite{xuanyuan} and FinEval~\cite{fineval}, each serving a unique purpose in our evaluation framework.

\noindent\textbf{FinanceIQ} is a Chinese evaluation dataset focused on the financial domain, designed to assess the knowledge and reasoning capabilities of LLMs in financial scenarios. It spans 10 broad financial categories and 36 subcategories, offering a total of 7,173 multiple-choice questions. This dataset is tailored to evaluate a wide range of financial knowledge, from basic concepts to complex problem-solving in areas such as accounting, banking, and financial analysis. 

\noindent\textbf{FinEval} is an expansive benchmark tailored for measuring the adeptness of LLMs in the finance domain. Encompassing 4,661 multiple-choice questions, it spans 34 academic subjects across the finance, economics, accounting, and professional certification domains. This dataset is structured to facilitate a layered evaluation through diverse assessment methods, including zero-shot, few-shot, answer-only, and chain-of-thought prompts.

\subsubsection{Legal Domain}
\textbf{LawBench~\cite{lawbench}:} 
For the legal domain, we implement the test over LawBench dataset. LawBench is a sophisticated benchmark designed to evaluate the legal knowledge capabilities of LLMs within the Chinese civil-law system. This benchmark tests LLMs across three cognitive levels: Memorization, Understanding, and Applying. It includes 20 tasks across five types: single-label classification, multi-label classification, regression, extraction, and generation. With over 500 cases, LawBench employs metrics including accuracy, F1, and ROUGE-L to ensure a comprehensive assessment of LLMs in diverse legal applications.


Through this diverse and challenging array of datasets, our model is subjected to an extensive evaluation, covering a range of tasks and scenarios in law and finance. This setup not only tests the model's domain-specific expertise but also its adaptability and versatility across different language settings.

\subsection{Comparison with other domain-specific models} 
For financial task, FinGPT~\cite{yang2023fingpt} and FinMA~\cite{xie2023pixiu} are adopted. Specifically, FinGPT v3.2, which uses LLaMA2-7B as a base model, and FinMA v0.1 NLP 7B version are adopted for a fair comparison.
For legal tasks, we conduct experiments compared with Hanfei~\cite{HanFei} and LaWGPT~\cite{LaWGPT}. Specifically, we chose the Legal-Base-7B version for LaWGPT, which is trained based on Chinese-LLaMA-7B~\cite{chinese-llama-alpaca}.

\begin{table}
	\centering
	\scriptsize
	\caption{The experimental results of general-domain and domain specialized LLMs on the financial FinEval~\cite{fineval} benchmark.}
	\begin{tabular}{cccccc}
		\toprule
		Model & Accounting & Certificate & Economy & Finance & \textbf{AVG} \\ 
		\midrule
		LLaMA2-7B         &   37.56          &     34.13          &   34.30       &  28.85 & 33.71   \\
		InternLM-7B       &  49.51          &     52.99          &   42.03       &  53.11 & 50.13   \\
		Baichuan2-7B      &  53.11          &     56.89          &   50.24       &  59.67 & 55.43   \\
		Qwen-7B           &  62.30          &     61.38         &   \textbf{56.04 }      &  \textbf{61.00} & 60.56   \\
		\midrule
		FinGPT-7B         & 29.18          &     25.45          &    30.90      &   25.90 & 27.54  \\
		FinMA-7B          & 35.08          &     32.34          &    34.78      &   35.41 &34.32  \\
		LawGPT-7B         & 25.25          &     31.74          &    28.50      &  27.54 & 28.32   \\
		HanFei-7B         & 29.84         &     29.64          &    27.54      &  31.48 &29.80   \\
		YunShan (Ours) & \textbf{65.57}          &     \textbf{ 69.46 }        &    51.21      &  55.08 & \textbf{61.34}  \\
		\bottomrule
	\end{tabular}
	\label{tab:experiment_finance2}
\end{table}

\subsubsection{Results on the Financial Benchmark}
\noindent\textbf{Results on FinancelQ. }
In Table~\ref{tab:experiment_finance1}, we first compare the general domain and domain- specialized LLMs on the FinancelQ benchmark. These models were assessed across multiple financial professional sub-domains, including Certified Public Accountant (CPA), Banking Qualification (BQ), Stock Portfolio Query (SPQ), Certified Financial Planner Qualification (CFPQ), Investment Portfolio Query (IPQ), Economist, Tax Accountant, Finance Qualification (FQ), Financial Planner, and Actuary. The "AVG" column represents the average score across all these sub-domains. The YunShan model demonstrated exceptional performance across nearly all sub-domains, particularly excelling in the Economist, Banking Qualification, and Certified Financial Planner Qualification categories, with scores of 72.50, 63.42, and 57.57, respectively. These scores significantly surpassed those of other models, highlighting its efficiency in handling finance-related tasks. Its average score of 54.56 was markedly higher than its counterparts, indicating its broad adaptability and proficiency in the financial domain. While models like Baichuan2-7B showed good performance in certain sub-domains such as Banking Qualification, Investment Portfolio Query, and Certified Financial Planner Qualification, they fell short in others. This suggests that while they are adaptable to specific financial sub-domains, they do not match the overall proficiency of the YunShan model. Conversely, other models such as FinGPT-7B and FinMA-7B exhibited generally weaker performance across all sub-domains, with average scores not exceeding 35. This may indicate a lesser degree of specialization in handling financial domain knowledge compared to other models. Overall, the exceptional performance of the YunShan model in this financial domain-specific evaluation reflects its outstanding ability in understanding and processing financial professional knowledge and showcases its potential as an LLM for the financial industry.

\noindent\textbf{Results on FinEval. } We also conducted an experiment on the FinEval benchmark in Table~\ref{tab:experiment_finance2}. The significant advantages of the YunShan model still exist for the FinEval benchmark. As shown in Table~\ref{tab:experiment_finance2}, models were assessed across four different sub-domains: Accounting, Certificate, Economy, and Finance, with the "AVG" column indicating the average score across all sub-domains. Specifically, we computed the average score by ${\rm AVG}=(305 \times {\rm Accounting} +334 \times {\rm Certificate}+207 \times {\rm Economy} +305 \times {\rm Finance})/1151$ according to the FinEval~\cite{fineval}. It's noteworthy that the YunShan model excelled in Certificate and Accounting, scoring 69.46 and 65.57 respectively, significantly outperforming other models. Its leading average score of 61.34 highlights its proficiency in handling finance-related tasks. Comparatively, Qwen-7B and Baichuan2-7B demonstrated robust performance, especially in the Finance sub-domain, with scores of 61.00 and 59.67 respectively. Their overall average scores, 60.56 and 55.43, indicate their strong adaptability to the financial sector. In contrast, models like FinGPT-7B, FinMA-7B, LawGPT-7B, and HanFei-7B showed relatively weaker performances across all sub-domains, with average scores below 35. This suggests that these models may be less specialized or inadequately trained for tasks specific to the financial domain. Overall, the YunShan model stands out as the most suitable model for financial domain tasks.

\begin{table}
	\centering
	\scriptsize
	\caption{The experimental results of general-domain and domain specialized LLMs on the legal benchmark.}
	\begin{tabular}{ccccc}
		\toprule
		Model & Memorization & Understanding & Applying & \textbf{AVG} \\ 
		\midrule
		LLaMA2-7B         &   1.52          &     9.09          &   16.19       &  11.16   \\
		InternLM-7B       &  3.10          &     8.29          &   29.76       &  16.36   \\
		Baichuan2-7B      &  2.78          &     8.13          &   19.79       &  12.26   \\
		Qwen-7B           &  0.82          &     17.43         &   31.54       &  21.41   \\
		\midrule
		FinGPT-7B         & 0.00          &     0.28          &    0.00      &   0.11  \\
		FinMA-7B          & 0.88          &     6.43          &    14.10      &   8.94  \\
		LawGPT-7B         & 1.69          &     3.41          &    19.64      &  9.73   \\
		HanFei-7B         & 13.71         &     6.91          &    25.22      &  14.92   \\
		YunShan (Ours) &\textbf{ 44.11 }         &     \textbf{ 19.72}         &    \textbf{43.70}      &   \textbf{31.75}  \\
		\bottomrule
	\end{tabular}
	\label{tab:experiment_law}
\end{table}

\subsubsection{Results on Legal Benchmark}
Table~\ref{tab:experiment_law} shows the overall zero-shot results of each model in the legal field on the different categories of LawBench~\cite{fei2023lawbench}, for three key aspects: Memorization, Understanding, and Applying. The average score is computed by ${\rm AVG}=(2\times{\rm Memorization} +10 \times {\rm Understanding}+8 \times {\rm Applying})/20$. Notably, the YunShan model yet again outperforms with its impressive scores, with the highest average score being 31.75. This performance indicates a superior capability in recalling legal information but also in understanding and applying it effectively, a crucial factor in legal contexts. Other models like Qwen-7B also demonstrate strong performances, with average scores of 21.35 and 21.41, respectively. In contrast, models such as FinGPT-7B exhibit significantly lower scores across all dimensions, leading to an overall average score of 0.11. Other models like InternLM-7B, Baichuan2-7B, and HanFei-7B demonstrate varied performances, with HanFei-7B scoring high in Memorization (13.71) but lower in Understanding (6.91), reflecting a potential imbalance in its skillset. Overall, the YunShan model demonstrates exceptional capability in the legal domain, outperforming others across all tested skills. This suggests that the YunShan model is adept at recalling legal information and excels in comprehending and applying it, making it a highly effective tool for legal applications.

\section{Conclusions and Discussions}
In this paper, we introduced PanGu-$\pi$, an innovative LLM architecture designed to mitigate the feature collapse issue in Transformer models by integrating nonlinearity. Our initial step involved a detailed examination of contemporary LLM architectures where we pinpointed the feature collapse challenge. Our theoretical analysis led us to propose the integration of nonlinearity, a principle predominantly used in convolutional neural networks for image processing, as a critical component for enhancing language models.	
To implement this, we augmented the FFN with a series informed activation function, thereby enhancing its nonlinearity. Additionally, we introduced an augmented shortcut to the MSA module, effectively reducing feature collapse. Theoretical evaluations revealed that these enhancements significantly boost the approximation capabilities of both the FFN and MSA modules. 
Leveraging these enhancements, we developed various sizes of PanGu-$\pi$, notably PanGu-$\pi$-7B and PanGu-$\pi$-1B. These models underwent extensive training and SFT, showcasing strong NLP capabilities. The PanGu-$\pi$, foundation models achieved state-of-the-art (SOTA) results in various downstream tasks, matching or surpassing similar-sized models. PanGu-$\pi$ displayed exceptional adaptability, delivering SOTA performances in the specialized domains of finance and law. 

\ifCLASSOPTIONcaptionsoff
\newpage
\fi

\renewcommand\refname{References}
{\small
	\bibliographystyle{unsrt2authabbrvpp}
	\bibliography{ref}

\begin{thebibliography}{100}

\bibitem{kaplan2020scaling}
J.~Kaplan et~al.
\newblock Scaling laws for neural language models.
\newblock {\em arXiv preprint arXiv:2001.08361}, 2020.

\bibitem{wei2022emergent}
J.~Wei et~al.
\newblock Emergent abilities of large language models.
\newblock {\em arXiv preprint arXiv:2206.07682}, 2022.

\bibitem{gpt3}
T.~B. Brown et~al.
\newblock Language models are few-shot learners.
\newblock In {\em NeurIPS}, 2020.

\bibitem{palm}
A.~Chowdhery et~al.
\newblock Palm: Scaling language modeling with pathways.
\newblock {\em arXiv preprint arXiv:2204.02311}, 2022.

\bibitem{gpt4}
OpenAI.
\newblock Gpt-4 technical report, 2023.

\bibitem{xi2023rise}
Z.~Xi et~al.
\newblock The rise and potential of large language model based agents: A
  survey.
\newblock {\em arXiv preprint arXiv:2309.07864}, 2023.

\bibitem{llama}
H.~Touvron et~al.
\newblock Llama: Open and efficient foundation language models.
\newblock {\em arXiv preprint arXiv:2302.13971}, 2023.

\bibitem{zeng2023glm-130b}
A.~Zeng et~al.
\newblock {GLM}-130b: An open bilingual pre-trained model.
\newblock In {\em The Eleventh International Conference on Learning
  Representations (ICLR)}, 2023.

\bibitem{yang2023baichuan}
A.~Yang et~al.
\newblock Baichuan 2: Open large-scale language models.
\newblock {\em arXiv preprint arXiv:2309.10305}, 2023.

\bibitem{LaWGPT}
S.~Pengxiao.
\newblock Lawgpt.
\newblock \url{https://github.com/pengxiao-song/LaWGPT}, 2023.

\bibitem{fingpt}
H.~Yang et~al.
\newblock Fingpt: Open-source financial large language models.
\newblock {\em FinLLM Symposium at IJCAI 2023}, 2023.

\bibitem{wang2023huatuo}
H.~Wang et~al.
\newblock Huatuo: Tuning llama model with chinese medical knowledge, 2023.

\bibitem{vaswani2017attention}
A.~Vaswani et~al.
\newblock Attention is all you need.
\newblock {\em Advances in neural information processing systems}, 30, 2017.

\bibitem{rwkv}
B.~Peng et~al.
\newblock Rwkv: Reinventing rnns for the transformer era.
\newblock {\em arXiv preprint arXiv:2305.13048}, 2023.

\bibitem{katharopoulos2020transformers}
A.~Katharopoulos et~al.
\newblock Transformers are rnns: Fast autoregressive transformers with linear
  attention.
\newblock In {\em International conference on machine learning}, 2020.

\bibitem{swiglu}
N.~Shazeer.
\newblock Glu variants improve transformer.
\newblock {\em arXiv preprint arXiv:2002.05202}, 2020.

\bibitem{fedus2022switch}
W.~Fedus et~al.
\newblock Switch transformers: Scaling to trillion parameter models with simple
  and efficient sparsity.
\newblock {\em The Journal of Machine Learning Research}, 23(1):5232--5270,
  2022.

\bibitem{augvit}
Y.~Tang et~al.
\newblock Augmented shortcuts for vision transformers.
\newblock In {\em NeurIPS}, volume~34, pp. 15316--15327, 2021.

\bibitem{dong2021attention}
Y.~Dong et~al.
\newblock Attention is not all you need: Pure attention loses rank doubly
  exponentially with depth.
\newblock In {\em ICML}, pp. 2793--2803. PMLR, 2021.

\bibitem{vanillanet}
H.~Chen et~al.
\newblock Vanillanet: the power of minimalism in deep learning.
\newblock In {\em NeurIPS}, 2023.

\bibitem{OpenAI-blog-2022-ChatGPT}
OpenAI.
\newblock Introducing chatgpt.
\newblock {\em OpenAI Blog}, November 2022.

\bibitem{OpenAI-OpenAI-2023-GPT-4}
OpenAI.
\newblock Gpt-4 technical report.
\newblock {\em OpenAI}, 2023.

\bibitem{touvron2023llama}
H.~Touvron et~al.
\newblock Llama: Open and efficient foundation language models.
\newblock {\em arXiv preprint arXiv:2302.13971}, 2023.

\bibitem{touvron2023llama2}
H.~Touvron et~al.
\newblock Llama 2: Open foundation and fine-tuned chat models.
\newblock {\em arXiv preprint arXiv:2307.09288}, 2023.

\bibitem{taori2023alpaca}
R.~Taori et~al.
\newblock Alpaca: A strong, replicable instruction-following model.
\newblock {\em Stanford Center for Research on Foundation Models. https://crfm.
  stanford. edu/2023/03/13/alpaca. html}, 3(6):7, 2023.

\bibitem{wang2022self}
Y.~Wang et~al.
\newblock Self-instruct: Aligning language model with self generated
  instructions.
\newblock {\em arXiv preprint arXiv:2212.10560}, 2022.

\bibitem{chiang2023vicuna}
W.-L. Chiang et~al.
\newblock Vicuna: An open-source chatbot impressing gpt-4 with 90\%* chatgpt
  quality.
\newblock {\em See https://vicuna. lmsys. org (accessed 14 April 2023)}, 2023.

\bibitem{Sharegpt}
D.~Eccleston.
\newblock Sharegpt.
\newblock {\em https://sharegpt.com/}, 2023.

\bibitem{team2023internlm}
I.~Team.
\newblock Internlm: A multilingual language model with progressively enhanced
  capabilities, 2023.

\bibitem{ren2023pangu}
X.~Ren et~al.
\newblock Pangu-$\{$$\backslash$Sigma$\}$: Towards trillion parameter language
  model with sparse heterogeneous computing.
\newblock {\em arXiv preprint arXiv:2303.10845}, 2023.

\bibitem{bai2023qwen}
J.~Bai et~al.
\newblock Qwen technical report.
\newblock {\em arXiv preprint arXiv:2309.16609}, 2023.

\bibitem{wei2023skywork}
T.~Wei et~al.
\newblock Skywork: A more open bilingual foundation model.
\newblock {\em arXiv preprint arXiv:2310.19341}, 2023.

\bibitem{zaheer2020big}
M.~Zaheer et~al.
\newblock Big bird: Transformers for longer sequences.
\newblock {\em Advances in neural information processing systems}, 2020.

\bibitem{child2019generating}
R.~Child et~al.
\newblock Generating long sequences with sparse transformers.
\newblock {\em arXiv preprint arXiv:1904.10509}, 2019.

\bibitem{roy2021efficient}
A.~Roy et~al.
\newblock Efficient content-based sparse attention with routing transformers.
\newblock {\em Transactions of the Association for Computational Linguistics},
  2021.

\bibitem{rae2019compressive}
J.~W. Rae et~al.
\newblock Compressive transformers for long-range sequence modelling.
\newblock {\em arXiv preprint arXiv:1911.05507}, 2019.

\bibitem{xiao2023efficient}
G.~Xiao et~al.
\newblock Efficient streaming language models with attention sinks.
\newblock {\em arXiv preprint arXiv:2309.17453}, 2023.

\bibitem{choromanski2020rethinking}
K.~Choromanski et~al.
\newblock Rethinking attention with performers.
\newblock {\em arXiv preprint arXiv:2009.14794}, 2020.

\bibitem{kitaev2020reformer}
N.~Kitaev et~al.
\newblock Reformer: The efficient transformer.
\newblock {\em arXiv preprint arXiv:2001.04451}, 2020.

\bibitem{beltagy2020longformer}
I.~Beltagy et~al.
\newblock Longformer: The long-document transformer.
\newblock {\em arXiv preprint arXiv:2004.05150}, 2020.

\bibitem{brown2020language}
T.~Brown et~al.
\newblock Language models are few-shot learners.
\newblock {\em Advances in neural information processing systems}, 2020.

\bibitem{dai2019transformer}
Z.~Dai et~al.
\newblock Transformer-xl: Attentive language models beyond a fixed-length
  context.
\newblock {\em arXiv preprint arXiv:1901.02860}, 2019.

\bibitem{martins2021infty}
P.~H. Martins et~al.
\newblock $\infty$-former: Infinite memory transformer.
\newblock {\em arXiv preprint arXiv:2109.00301}, 2021.

\bibitem{sun2023retentive}
Y.~Sun et~al.
\newblock Retentive network: A successor to transformer for large language
  models.
\newblock {\em arXiv preprint arXiv:2307.08621}, 2023.

\bibitem{poli2023hyena}
M.~Poli et~al.
\newblock Hyena hierarchy: Towards larger convolutional language models.
\newblock {\em arXiv preprint arXiv:2302.10866}, 2023.

\bibitem{peng2023rwkv}
B.~Peng et~al.
\newblock Rwkv: Reinventing rnns for the transformer era.
\newblock {\em arXiv preprint arXiv:2305.13048}, 2023.

\bibitem{du2022glam}
N.~Du et~al.
\newblock Glam: Efficient scaling of language models with mixture-of-experts.
\newblock In {\em International Conference on Machine Learning}, 2022.

\bibitem{roller2021hash}
S.~Roller et~al.
\newblock Hash layers for large sparse models.
\newblock {\em Advances in Neural Information Processing Systems}, 2021.

\bibitem{chi2022representation}
Z.~Chi et~al.
\newblock On the representation collapse of sparse mixture of experts.
\newblock {\em Advances in Neural Information Processing Systems}, 2022.

\bibitem{lewis2021base}
M.~Lewis et~al.
\newblock Base layers: Simplifying training of large, sparse models.
\newblock In {\em International Conference on Machine Learning}, 2021.

\bibitem{chowdhery2022palm}
A.~Chowdhery et~al.
\newblock Palm: Scaling language modeling with pathways.
\newblock {\em arXiv preprint arXiv:2204.02311}, 2022.

\bibitem{shazeer2020glu}
N.~Shazeer.
\newblock Glu variants improve transformer.
\newblock {\em arXiv preprint arXiv:2002.05202}, 2020.

\bibitem{wang2021pyramid}
W.~Wang et~al.
\newblock Pyramid vision transformer: A versatile backbone for dense prediction
  without convolutions.
\newblock In {\em Proceedings of the IEEE/CVF international conference on
  computer vision}, 2021.

\bibitem{liu2021swin}
Z.~Liu et~al.
\newblock Swin transformer: Hierarchical vision transformer using shifted
  windows.
\newblock In {\em Proceedings of the IEEE/CVF international conference on
  computer vision}, 2021.

\bibitem{dosovitskiy2020image}
A.~Dosovitskiy et~al.
\newblock An image is worth 16x16 words: Transformers for image recognition at
  scale.
\newblock {\em arXiv preprint arXiv:2010.11929}, 2020.

\bibitem{guo2022cmt}
J.~Guo et~al.
\newblock Cmt: Convolutional neural networks meet vision transformers.
\newblock In {\em Proceedings of the IEEE/CVF Conference on Computer Vision and
  Pattern Recognition}, 2022.

\bibitem{heo2021rethinking}
B.~Heo et~al.
\newblock Rethinking spatial dimensions of vision transformers.
\newblock In {\em Proceedings of the IEEE/CVF International Conference on
  Computer Vision}, 2021.

\bibitem{pan2021scalable}
Z.~Pan et~al.
\newblock Scalable vision transformers with hierarchical pooling.
\newblock In {\em Proceedings of the IEEE/CVF International Conference on
  Computer Vision}, 2021.

\bibitem{chen2021crossvit}
C.-F.~R. Chen et~al.
\newblock Crossvit: Cross-attention multi-scale vision transformer for image
  classification.
\newblock In {\em Proceedings of the IEEE/CVF international conference on
  computer vision}, 2021.

\bibitem{graham2021levit}
B.~Graham et~al.
\newblock Levit: a vision transformer in convnet's clothing for faster
  inference.
\newblock In {\em Proceedings of the IEEE/CVF international conference on
  computer vision}, 2021.

\bibitem{mehta2021mobilevit}
S.~Mehta and M.~Rastegari.
\newblock Mobilevit: light-weight, general-purpose, and mobile-friendly vision
  transformer.
\newblock {\em arXiv preprint arXiv:2110.02178}, 2021.

\bibitem{han2021transformer}
K.~Han et~al.
\newblock Transformer in transformer.
\newblock {\em Advances in Neural Information Processing Systems}, 2021.

\bibitem{parmar2018image}
N.~Parmar et~al.
\newblock Image transformer.
\newblock In {\em International conference on machine learning}, 2018.

\bibitem{liu2023efficientvit}
X.~Liu et~al.
\newblock Efficientvit: Memory efficient vision transformer with cascaded group
  attention.
\newblock In {\em Proceedings of the IEEE/CVF Conference on Computer Vision and
  Pattern Recognition}, 2023.

\bibitem{xiong2021nystromformer}
Y.~Xiong et~al.
\newblock Nystr{\"o}mformer: A nystr{\"o}m-based algorithm for approximating
  self-attention.
\newblock In {\em Proceedings of the AAAI Conference on Artificial
  Intelligence}, 2021.

\bibitem{guo2022beyond}
M.-H. Guo et~al.
\newblock Beyond self-attention: External attention using two linear layers for
  visual tasks.
\newblock {\em IEEE Transactions on Pattern Analysis and Machine Intelligence},
  2022.

\bibitem{guo2022hire}
J.~Guo et~al.
\newblock Hire-mlp: Vision mlp via hierarchical rearrangement.
\newblock In {\em Proceedings of the IEEE/CVF Conference on Computer Vision and
  Pattern Recognition}, 2022.

\bibitem{tang2022image}
Y.~Tang et~al.
\newblock An image patch is a wave: Phase-aware vision mlp.
\newblock In {\em Proceedings of the IEEE/CVF Conference on Computer Vision and
  Pattern Recognition}, 2022.

\bibitem{yu2022metaformer}
W.~Yu et~al.
\newblock Metaformer is actually what you need for vision.
\newblock In {\em Proceedings of the IEEE/CVF conference on computer vision and
  pattern recognition}, 2022.

\bibitem{lian2021mlp}
D.~Lian et~al.
\newblock As-mlp: An axial shifted mlp architecture for vision.
\newblock {\em arXiv preprint arXiv:2107.08391}, 2021.

\bibitem{chen2021cyclemlp}
S.~Chen et~al.
\newblock Cyclemlp: A mlp-like architecture for dense prediction.
\newblock {\em arXiv preprint arXiv:2107.10224}, 2021.

\bibitem{yuan2021tokens}
L.~Yuan et~al.
\newblock Tokens-to-token vit: Training vision transformers from scratch on
  imagenet.
\newblock In {\em Proceedings of the IEEE/CVF international conference on
  computer vision}, 2021.

\bibitem{dong2022cswin}
X.~Dong et~al.
\newblock Cswin transformer: A general vision transformer backbone with
  cross-shaped windows.
\newblock In {\em Proceedings of the IEEE/CVF Conference on Computer Vision and
  Pattern Recognition}, 2022.

\bibitem{rao2021global}
Y.~Rao et~al.
\newblock Global filter networks for image classification.
\newblock {\em Advances in neural information processing systems}, 2021.

\bibitem{touvron2022resmlp}
H.~Touvron et~al.
\newblock Resmlp: Feedforward networks for image classification with
  data-efficient training.
\newblock {\em IEEE Transactions on Pattern Analysis and Machine Intelligence},
  2022.

\bibitem{tolstikhin2021mlp}
I.~O. Tolstikhin et~al.
\newblock Mlp-mixer: An all-mlp architecture for vision.
\newblock {\em Advances in neural information processing systems}, 2021.

\bibitem{wu2023bloomberggpt}
S.~Wu et~al.
\newblock Bloomberggpt: A large language model for finance.
\newblock {\em arXiv preprint arXiv:2303.17564}, 2023.

\bibitem{workshop2022bloom}
B.~Workshop et~al.
\newblock Bloom: A 176b-parameter open-access multilingual language model.
\newblock {\em arXiv preprint arXiv:2211.05100}, 2022.

\bibitem{yang2023fingpt}
H.~Yang et~al.
\newblock Fingpt: Open-source financial large language models.
\newblock {\em arXiv preprint arXiv:2306.06031}, 2023.

\bibitem{liu2023fingpt}
X.-Y. Liu et~al.
\newblock Fingpt: Democratizing internet-scale data for financial large
  language models.
\newblock {\em arXiv preprint arXiv:2307.10485}, 2023.

\bibitem{xie2023pixiu}
Q.~Xie et~al.
\newblock Pixiu: A large language model, instruction data and evaluation
  benchmark for finance.
\newblock {\em arXiv preprint arXiv:2306.05443}, 2023.

\bibitem{shah2022flue}
R.~S. Shah et~al.
\newblock When flue meets flang: Benchmarks and large pre-trained language
  model for financial domain.
\newblock {\em arXiv preprint arXiv:2211.00083}, 2022.

\bibitem{zhang2023xuanyuan}
X.~Zhang and Q.~Yang.
\newblock Xuanyuan 2.0: A large chinese financial chat model with hundreds of
  billions parameters.
\newblock In {\em Proceedings of the 32nd ACM International Conference on
  Information and Knowledge Management}, pp. 4435--4439, 2023.

\bibitem{chen2023disc}
W.~Chen et~al.
\newblock Disc-finllm: A chinese financial large language model based on
  multiple experts fine-tuning.
\newblock {\em arXiv preprint arXiv:2310.15205}, 2023.

\bibitem{Baichuan13B}
Baichuan-inc.
\newblock Baichuan-13b.
\newblock \url{https://github.com/baichuan-inc/Baichuan-13B}, 2023.

\bibitem{BaoLuo-LawAssistant}
xuanxuanzl.
\newblock Baoluo lawassistant.
\newblock \url{https://github.com/xuanxuanzl/BaoLuo-LawAssistant}, 2023.

\bibitem{Lychee}
davidpig.
\newblock Lychee.
\newblock \url{https://github.com/davidpig/lychee_law}, 2023.

\bibitem{JurisLMs}
seudl.
\newblock Jurislms: Jurisprudential language models.
\newblock \url{https://github.com/seudl/JurisLMs}, 2023.

\bibitem{yue2023fedjudge}
L.~Yue et~al.
\newblock Fedjudge: Federated legal large language model.
\newblock {\em arXiv preprint arXiv:2309.08173}, 2023.

\bibitem{mcmahan2017communication}
B.~McMahan et~al.
\newblock Communication-efficient learning of deep networks from decentralized
  data.
\newblock In {\em Artificial intelligence and statistics}, pp. 1273--1282.
  PMLR, 2017.

\bibitem{zhou2021image}
M.~Zhou et~al.
\newblock Image de-raining via continual learning.
\newblock In {\em Proceedings of the IEEE/CVF Conference on Computer Vision and
  Pattern Recognition}, pp. 4907--4916, 2021.

\bibitem{HanFei}
W.~He et~al.
\newblock Hanfei-1.0.
\newblock \url{https://github.com/siat-nlp/HanFei}, 2023.

\bibitem{huang2023lawyer}
Q.~Huang et~al.
\newblock Lawyer llama technical report.
\newblock {\em arXiv preprint arXiv:2305.15062}, 2023.

\bibitem{wisdomInterrogatory}
zhihaiLLM.
\newblock wisdominterrogatory.
\newblock \url{https://github.com/zhihaiLLM/wisdomInterrogatory}, 2023.

\bibitem{fuzi.mingcha}
S.~Wu et~al.
\newblock fuzi.mingcha.
\newblock \url{https://github.com/irlab-sdu/fuzi.mingcha}, 2023.

\bibitem{LAWGPT-zh}
H.~Liu et~al.
\newblock Lawgpt.
\newblock \url{https://github.com/LiuHC0428/LAW_GPT}, 2023.

\bibitem{LexiLaw}
CSHaitao.
\newblock Lexilaw.
\newblock \url{https://github.com/CSHaitao/LexiLaw}, 2023.

\bibitem{cui2023chatlaw}
J.~Cui et~al.
\newblock Chatlaw: Open-source legal large language model with integrated
  external knowledge bases.
\newblock {\em arXiv preprint arXiv:2306.16092}, 2023.

\bibitem{yue2023disclawllm}
S.~Yue et~al.
\newblock Disc-lawllm: Fine-tuning large language models for intelligent legal
  services, 2023.

\bibitem{shi2022revisiting}
H.~Shi et~al.
\newblock Revisiting over-smoothing in bert from the perspective of graph.
\newblock {\em arXiv preprint arXiv:2202.08625}, 2022.

\bibitem{he2015delving}
K.~He et~al.
\newblock Delving deep into rectifiers: Surpassing human-level performance on
  imagenet classification.
\newblock In {\em Proceedings of the IEEE international conference on computer
  vision}, pp. 1026--1034, 2015.

\bibitem{hendrycks2016gaussian}
D.~Hendrycks and K.~Gimpel.
\newblock Gaussian error linear units (gelus).
\newblock {\em arXiv preprint arXiv:1606.08415}, 2016.

\bibitem{ramachandran2017searching}
P.~Ramachandran et~al.
\newblock Searching for activation functions.
\newblock {\em arXiv preprint arXiv:1710.05941}, 2017.

\bibitem{tang2021augmented}
Y.~Tang et~al.
\newblock Augmented shortcuts for vision transformers.
\newblock {\em Advances in Neural Information Processing Systems},
  34:15316--15327, 2021.

\bibitem{shibata1999byte}
Y.~Shibata et~al.
\newblock Byte pair encoding: A text compression scheme that accelerates
  pattern matching.
\newblock {\em Technical Report DOI-TR-161, Department of Informatics, Kyushu
  University}, 1999.

\bibitem{kudo2018sentencepiece}
T.~Kudo and J.~Richardson.
\newblock Sentencepiece: A simple and language independent subword tokenizer
  and detokenizer for neural text processing.
\newblock {\em arXiv preprint arXiv:1808.06226}, 2018.

\bibitem{loshchilov2017decoupled}
I.~Loshchilov and F.~Hutter.
\newblock Decoupled weight decay regularization.
\newblock {\em arXiv preprint arXiv:1711.05101}, 2017.

\bibitem{loshchilov2016sgdr}
I.~Loshchilov and F.~Hutter.
\newblock Sgdr: Stochastic gradient descent with warm restarts.
\newblock {\em arXiv preprint arXiv:1608.03983}, 2016.

\bibitem{zhang2019root}
B.~Zhang and R.~Sennrich.
\newblock Root mean square layer normalization.
\newblock {\em Advances in Neural Information Processing Systems}, 32, 2019.

\bibitem{su2023roformer}
J.~Su et~al.
\newblock Roformer: Enhanced transformer with rotary position embedding.
\newblock {\em Neurocomputing}, pp. 127063, 2023.

\bibitem{2023opencompass}
O.~Contributors.
\newblock Opencompass: A universal evaluation platform for foundation models.
\newblock \url{https://github.com/open-compass/opencompass}, 2023.

\bibitem{huang2023c}
Y.~Huang et~al.
\newblock C-eval: A multi-level multi-discipline chinese evaluation suite for
  foundation models.
\newblock {\em arXiv preprint arXiv:2305.08322}, 2023.

\bibitem{li2023cmmlu}
H.~Li et~al.
\newblock Cmmlu: Measuring massive multitask language understanding in chinese.
\newblock {\em arXiv preprint arXiv:2306.09212}, 2023.

\bibitem{hendrycks2020measuring}
D.~Hendrycks et~al.
\newblock Measuring massive multitask language understanding.
\newblock {\em arXiv preprint arXiv:2009.03300}, 2020.

\bibitem{zhong2023agieval}
W.~Zhong et~al.
\newblock Agieval: A human-centric benchmark for evaluating foundation models,
  2023.

\bibitem{clark2019boolq}
C.~Clark et~al.
\newblock Boolq: Exploring the surprising difficulty of natural yes/no
  questions.
\newblock {\em arXiv preprint arXiv:1905.10044}, 2019.

\bibitem{wang2020superglue}
A.~Wang et~al.
\newblock Superglue: A stickier benchmark for general-purpose language
  understanding systems, 2020.

\bibitem{bisk2019piqa}
Y.~Bisk et~al.
\newblock Piqa: Reasoning about physical commonsense in natural language, 2019.

\bibitem{li2022csl}
Y.~Li et~al.
\newblock Csl: A large-scale chinese scientific literature dataset.
\newblock {\em arXiv preprint arXiv:2209.05034}, 2022.

\bibitem{xu2021fewclue}
L.~Xu et~al.
\newblock Fewclue: A chinese few-shot learning evaluation benchmark.
\newblock {\em arXiv preprint arXiv:2107.07498}, 2021.

\bibitem{narayan2018dont}
S.~Narayan et~al.
\newblock Don't give me the details, just the summary! topic-aware
  convolutional neural networks for extreme summarization, 2018.

\bibitem{hu2016lcsts}
B.~Hu et~al.
\newblock Lcsts: A large scale chinese short text summarization dataset, 2016.

\bibitem{xue2022go}
F.~Xue et~al.
\newblock Go wider instead of deeper.
\newblock In {\em Proceedings of the AAAI Conference on Artificial
  Intelligence}, volume~36, pp. 8779--8787, 2022.

\bibitem{Cai2021IsotropyIT}
X.~Cai et~al.
\newblock Isotropy in the contextual embedding space: Clusters and manifolds.
\newblock In {\em International Conference on Learning Representations}, 2021.

\bibitem{Ethayarajh2019HowCA}
K.~Ethayarajh.
\newblock How contextual are contextualized word representations? comparing the
  geometry of bert, elmo, and gpt-2 embeddings.
\newblock In {\em Conference on Empirical Methods in Natural Language
  Processing}, 2019.

\bibitem{2002Building}
M.~P. Marcus et~al.
\newblock Building a large annotated corpus of english: The penn treebank.
\newblock {\em Computational Linguistics}, 19:313--330, 2002.

\bibitem{2021SimCSE}
T.~Gao et~al.
\newblock Simcse: Simple contrastive learning of sentence embeddings.
\newblock 2021.

\bibitem{chinese-llama-alpaca}
Y.~Cui et~al.
\newblock Efficient and effective text encoding for chinese llama and alpaca.
\newblock {\em arXiv preprint arXiv:2304.08177}, 2023.

\bibitem{tinyllama}
T.~W. Peiyuan~Zhang, Guangtao~Zeng and W.~Lu.
\newblock Tinyllama, 2023.

\bibitem{xia2023sheared}
M.~Xia et~al.
\newblock Sheared llama: Accelerating language model pre-training via
  structured pruning.
\newblock {\em arXiv preprint arXiv:2310.06694}, 2023.

\bibitem{2022pileoflaw}
P.~Henderson et~al.
\newblock Pile of law: Learning responsible data filtering from the law and a
  256gb open-source legal dataset, 2022.

\bibitem{2023-lexfiles}
I.~Chalkidis et~al.
\newblock {L}e{XF}iles and {L}egal{LAMA}: Facilitating {E}nglish multinational
  legal language model development.
\newblock In {\em Proceedings of the 61st Annual Meeting of the Association for
  Computational Linguistics (Volume 1: Long Papers)}, pp. 15513--15535.
  Association for Computational Linguistics, July 2023.

\bibitem{zhong2019jec}
H.~Zhong et~al.
\newblock J{EC-QA}: A legal-domain question answering dataset.
\newblock In {\em Proceedings of AAAI}, 2020.

\bibitem{wang2020neural}
C.~Wang et~al.
\newblock Neural machine translation with byte-level subwords.
\newblock In {\em Proceedings of the AAAI conference on artificial
  intelligence}, volume~34, pp. 9154--9160, 2020.

\bibitem{zhong2020jec}
H.~Zhong et~al.
\newblock Jec-qa: a legal-domain question answering dataset.
\newblock In {\em Proceedings of the AAAI Conference on Artificial
  Intelligence}, volume~34, pp. 9701--9708, 2020.

\bibitem{xuanyuan}
X.~Zhang and Q.~Yang.
\newblock Xuanyuan 2.0: A large chinese financial chat model with hundreds of
  billions parameters.
\newblock In {\em Proceedings of the 32nd ACM International Conference on
  Information and Knowledge Management}, pp. 4435--4439, 2023.

\bibitem{fineval}
L.~Zhang et~al.
\newblock Fineval: A chinese financial domain knowledge evaluation benchmark
  for large language models, 2023.

\bibitem{lawbench}
Z.~Fei et~al.
\newblock Lawbench: Benchmarking legal knowledge of large language models.
\newblock {\em arXiv preprint arXiv:2309.16289}, 2023.

\bibitem{fei2023lawbench}
Z.~Fei et~al.
\newblock Lawbench: Benchmarking legal knowledge of large language models.
\newblock {\em arXiv preprint arXiv:2309.16289}, 2023.

\end{thebibliography}
}

\newpage

\newpage

\appendix

\textbf{Proof of Lemma~\ref{lm1}}

\begin{proof}
	
	Write $\boldsymbol{WW}^\top=\boldsymbol{PDP}^\top$ for the eigin-decomposition of $\boldsymbol{WW}^\top$, where $\boldsymbol{P}=[\boldsymbol{p}_1,\boldsymbol{p}_2, \dots, \boldsymbol{p}_d]$ is the standard orthogonal and $\boldsymbol{D}=\text{diag}(d_1, \dots, d_d)$ with all $d_1 \geq \dots \geq d_d\geq 0$ and for nonzero eigenvalues $d_i = s_i^2 > 0$, where $s_i$ is the i-th largest singular value of $\boldsymbol{W}$.
	
	$\boldsymbol{HH}^\top=\boldsymbol{Q\Omega Q}^\top$ for the eigin-decomposition of $\boldsymbol{HH}^\top$, where $\boldsymbol{Q}=[\boldsymbol{q}_1,\boldsymbol{q}_2, \dots, \boldsymbol{q}_N]$ is the orthogonal and $\boldsymbol{\Omega}=\text{diag}(\omega_1, \dots, \omega_N)$ with all $\omega_i\geq 0$. And $\boldsymbol{e}=N^{-1/2}[1,1, \dots, 1]^\top= N^{-1/2} \boldsymbol{1} \in\mathbb{R}^{N\times 1}$.
	
	(1) The formula proves as follows:
	\begin{align*}
		d_{\mathcal{M}_m}(\boldsymbol{H}\boldsymbol{W})^2
		&=\Vert(\boldsymbol{I}-\boldsymbol{ee}^\top)\boldsymbol{H}\boldsymbol{W} \Vert^2_F \\
		&=tr\{(\boldsymbol{I}-\boldsymbol{ee}^\top)\boldsymbol{H}\boldsymbol{WW}^\top\boldsymbol{H}^\top(\boldsymbol{I}-\boldsymbol{ee}^\top)\} \\
		&=tr\{\boldsymbol{WW}^\top\boldsymbol{H}^\top(\boldsymbol{I}-\boldsymbol{ee}^\top)\boldsymbol{H}\}\\
		&=tr\{\boldsymbol{PD{P}^\top}\boldsymbol{H}^\top(\boldsymbol{I}-\boldsymbol{ee}^\top)\boldsymbol{H}\}\\
		&=tr\{\boldsymbol{D{P}^\top}\boldsymbol{H}^\top(\boldsymbol{I}-\boldsymbol{ee}^\top)\boldsymbol{HP}\}\\
		&=tr\{\boldsymbol{D{P}^\top}\boldsymbol{H}^\top(\boldsymbol{I}-\boldsymbol{ee}^\top)\boldsymbol{HP}\}\\
		&=\sum_{i=1}^N d_i\boldsymbol{p}_i^\top\boldsymbol{H}^\top(\boldsymbol{I}-\boldsymbol{ee}^\top)\boldsymbol{H}\boldsymbol{p}_i\\
		&\leq \sum_{i=1}^N s^2\boldsymbol{p}_i^\top\boldsymbol{H}^\top(\boldsymbol{I}-\boldsymbol{ee}^\top)\boldsymbol{H}\boldsymbol{p}_i\\
		&=s^2d_{\mathcal{M}_d}(\boldsymbol{H})^2.
	\end{align*}
	Since matrix $\boldsymbol{H}^\top(\boldsymbol{I}-\boldsymbol{ee}^\top)\boldsymbol{H}$ is positive semidefinite, $\boldsymbol{p}_i^\top\boldsymbol{H}^\top(\boldsymbol{I}-\boldsymbol{ee}^\top)\boldsymbol{H}\boldsymbol{p}_i \geq 0$.
	
	It follows that $d_{\mathcal{M}_m}(\boldsymbol{H}\boldsymbol{W}) \leq sd_{\mathcal{M}_d}(\boldsymbol{H})$.

	Note that $d_{\mathcal{M}_d}(\boldsymbol{H}) = \Vert \boldsymbol{H}-\boldsymbol{1x}_{\min}^\top\Vert_F$, where $\boldsymbol{x}_{\min}^\top = \arg\min_{\boldsymbol{x}_l} \Vert \boldsymbol{H}-\boldsymbol{1x}_{l}^\top \Vert_F$. And $\Vert \sigma(\boldsymbol{H}_1)-\sigma(\boldsymbol{H}_2)\Vert_F \leq L \Vert \boldsymbol{H}_1 -\boldsymbol{H}_2\Vert_F $
	
	(2) The formula proves as follows:
	\begin{align*}
		d_{\mathcal{M}_d}(\sigma(\boldsymbol{H})) &= \Vert \sigma(\boldsymbol{H})-\boldsymbol{1{x^{\sigma}}_{\min}}^\top\Vert_F \\
		&\leq \Vert \sigma(\boldsymbol{H})-\boldsymbol{1}{\sigma(\boldsymbol{x}_{\min})}^\top\Vert_F \\
		&=  \Vert \sigma(\boldsymbol{H})-\sigma(\boldsymbol{1}\boldsymbol{x}_{\min}^\top)\Vert_F \\
		&\leq  L \Vert \boldsymbol{H} -\boldsymbol{1}\boldsymbol{x}_{\min}^\top\Vert_F \\
		&= L d_{\mathcal{M}_d}(\boldsymbol{H}), \\
	\end{align*}

	(3) The formula proves as follows:
	\begin{align*}
		&\alpha_1 d_{\mathcal{M}_d}(\boldsymbol{H}) + \alpha_2 d_{\mathcal{M}_d}(\boldsymbol{B}) \\
		&=  \alpha_1 \Vert \boldsymbol{H}-\boldsymbol{1{x^{H}_{\min}}^\top}\Vert_F + \alpha_2 \Vert \boldsymbol{B}-\boldsymbol{1{x^{B}_{\min}}^\top}\Vert_F\\
		&\geq  \Vert \alpha_1\boldsymbol{H} + \alpha_2  \boldsymbol{B} - \boldsymbol{1{(\alpha_1x^{H}_{\min}+\alpha_2x^{B}_{\min})}^\top} \Vert_F \\
		&\geq  \Vert \alpha_1\boldsymbol{H} + \alpha_2  \boldsymbol{B} - \boldsymbol{1{(x^{\alpha_1H+\alpha_2B}_{\min})}^\top} \Vert_F \\
		&= d_{\mathcal{M}_d}(\alpha_1 \boldsymbol{H}+\alpha_2 \boldsymbol{B})\\
	\end{align*}

	For the last inequality, we refer to \cite{shi2022revisiting}.
\end{proof}

\textbf{Proof of Theorem~\ref{th1-plain}}

\begin{proof}
	
	\begin{align*}
		d_{\mathcal{M}_m}(\boldsymbol{AZ}_l\boldsymbol{W}) 
		&\leq \sqrt{\lambda}d_{\mathcal{M}_m}(\boldsymbol{Z}_{l}\boldsymbol{W}) \\
		&\leq \sqrt{\lambda}sd_{\mathcal{M}_m}(\boldsymbol{Z_{l}}) \\
	\end{align*}

\end{proof}

\textbf{Proof of Lemma~\ref{lm2}}

\begin{proof}
	
	\begin{align*}
		&d_{\mathcal{M}_{Hm}}(\Concat ([\boldsymbol{H}_h]_{h=1}^H))^2 \\
		=&\Vert(\boldsymbol{I}-\boldsymbol{ee}^\top)\Concat ([\boldsymbol{H}_h]_{h=1}^H) \Vert^2_F \\
		=&{tr\{(\boldsymbol{I}-\boldsymbol{ee}^\top)\Concat ([\boldsymbol{H}_h]_{h=1}^H)\Concat ([\boldsymbol{H}_h]_{h=1}^H)^\top\}} \\
		=&tr\{\Concat ([\boldsymbol{H}_h]_{h=1}^H)\Concat ([\boldsymbol{H}_h]_{h=1}^H)^\top\}\\
		&-tr\{\Concat ([\boldsymbol{e}^\top\boldsymbol{H}_h]_{h=1}^H)\Concat ([\boldsymbol{ee}^\top\boldsymbol{H}_h]_{h=1}^H)^\top\} \\
		=&\sum_{h=1}^{H}{tr\{(\boldsymbol{I}-\boldsymbol{ee}^\top)\boldsymbol{H}_h\boldsymbol{H}_h^\top\}} \\
		=& \sum_{h=1}^{H}d_{\mathcal{M}_m}(\boldsymbol{H}_h)^2,
	\end{align*}

\end{proof}

\textbf{Proof of Theorem~\ref{th2-plain}}

\begin{proof}

	\begin{align*}
		{\MSA} (\boldsymbol{Z}_l)=  \Concat ([\boldsymbol{A}_{lh}\boldsymbol{Z}_{l}\boldsymbol{W}^v_{lh}]_{h=1}^H)\boldsymbol{W}^o_{l},
	\end{align*}

	\begin{align*}
		d_{\mathcal{M}_d}(\boldsymbol{Z}_{l+1})&= d_{\mathcal{M}_d}(\Concat ([\boldsymbol{A}_{lh}\boldsymbol{Z}_{l}\boldsymbol{W}^v_{lh}]_{h=1}^H)\boldsymbol{W}^o_{l}) \\
		&\leq \upsilon_1 d_{\mathcal{M}_d}(\Concat ([\boldsymbol{A}_{lh}\boldsymbol{Z}_{l}\boldsymbol{W}^v_{lh}]_{h=1}^H)) \\
		&\leq \upsilon_1\sqrt{\sum_{h=1}^H d_{\mathcal{M}_d}(\boldsymbol{A}_{lh}\boldsymbol{Z}_{l}\boldsymbol{W}^v_{lh})^2}\\ 
		&\leq \sqrt{\lambda}s\upsilon_1\sqrt{\sum_{h=1}^H d_{\mathcal{M}_d}(\boldsymbol{Z}_{l})^2} \\
		&= \sqrt{\lambda H}s\upsilon_1d_{\mathcal{M}_d}(\boldsymbol{Z}_{l}) \\
		&\leq (\sqrt{\lambda H} s\upsilon_1)^{l+1}d_{\mathcal{M}_d}(\boldsymbol{Z}_0). \\
	\end{align*}
	
\end{proof}





%

\textbf{Proof of Theorem~\ref{th3-mlp}}

\begin{proof}
	
	\begin{align*}
		\MLP (\boldsymbol{Z}'_{l}) = \sigma(\boldsymbol{Z}'_{l}\boldsymbol{W}'_{l_1})\boldsymbol{W}'_{l_2},~l \in [1,2,\cdots, L],
	\end{align*}
	
	\begin{align*}
		d_{\mathcal{M}_d}(\MLP (\boldsymbol{Z}'_{l})))&= d_{\mathcal{M}_d}(\sigma(\boldsymbol{Z}'_{l}\boldsymbol{W}'_{l_1})\boldsymbol{W}'_{l_2}) \\
		&\leq \upsilon_2d_{\mathcal{M}_d}(\sigma(\boldsymbol{Z}'_{l}\boldsymbol{W}'_{l_1})) \\
		&\leq L\upsilon_2d_{\mathcal{M}_d}(\boldsymbol{Z}'_{l}\boldsymbol{W}'_{l_1}) \\ 
		&\leq  Ls\upsilon_2d_{\mathcal{M}_d}(\boldsymbol{Z}'_{l}) \\
	\end{align*}
	
	\begin{align*}
		d_{\mathcal{M}_d}(\boldsymbol{Z}'_l)&\leq (Ls\upsilon_2)d_{\mathcal{M}_d}(\boldsymbol{Z}'_{l-1}) \\
		&\leq (Ls\upsilon_2)^{l}d_{\mathcal{M}_d}(\boldsymbol{Z}'_0).
	\end{align*}
	
\end{proof}

\textbf{Proof of Theorem~\ref{th5-aug}}

\begin{proof}
	\begin{align*}
		{\rm AugMSA} (\boldsymbol{Z}_{l})=\MSA(\boldsymbol{Z}_l) + \boldsymbol{Z}_l  +  \sum_{i=1
		}^T \T_{{li}}(\boldsymbol{Z}_{l};\boldsymbol{\Theta}_{li}),\\ 
	\end{align*}
	
	\begin{align*}
		&d_{\mathcal{M}_d}({\rm AugMSA} (\boldsymbol{Z_{l}}))\\
		&= d_{\mathcal{M}_d}(\MSA(\boldsymbol{Z}_l) + \boldsymbol{Z}_l  +  \sum_{i=1
		}^T \T_{{li}}(\boldsymbol{Z}_{l};\boldsymbol{\Theta}_{li})) \\
		&\leq d_{\mathcal{M}_d}(\MSA(\boldsymbol{Z}_l)) + d_{\mathcal{M}_d}(\boldsymbol{Z}_l) + d_{\mathcal{M}_d}(\sum_{i=1}^T \T_{{li}}(\boldsymbol{Z}_{l};\boldsymbol{\Theta}_{li}))\\
		&\leq d_{\mathcal{M}_d}(\MSA(\boldsymbol{Z}_l)) + d_{\mathcal{M}_d}(\boldsymbol{Z}_l) + \sum_{i=1}^T d_{\mathcal{M}_d}(\sigma(\boldsymbol{Z}_{l}\boldsymbol{\Theta}_{li})) \\   
		&\leq  (\sqrt{\lambda H} s\upsilon_1 + 1)d_{\mathcal{M}_d}(\boldsymbol{Z}'_{l}) + L\sum_{i=1}^T d_{\mathcal{M}_d}(\boldsymbol{Z}_{l}\boldsymbol{\Theta}_{li})\\
		&\leq (\sqrt{\lambda H} s\upsilon_1 + 1+ \sum_{i=1}^T L\Vert \boldsymbol{\Theta}_{li}\Vert_2)d_{\mathcal{M}_d}(\boldsymbol{Z}_l).
	\end{align*}
	
	Considering that H heads exist inthe MSA module, the diversity $d_\mathcal{M}(\boldsymbol{Z}_{l+1})$ becomes:
	
	\begin{align*}
		&d_{\mathcal{M}_d}(\boldsymbol{Z}_{l})
		\leq (\sqrt{\lambda H} s\upsilon_1 + 1+ \sum_{i=1}^T L\Vert \boldsymbol{\Theta}_{li}\Vert_2)^ld_{\mathcal{M}_d}(\boldsymbol{Z}_0).
	\end{align*}
\end{proof}

\textbf{Proof of Lemma 3}


\begin{proof}  
	When $H=1$,
	\begin{align*}
		{\MSA} (\boldsymbol{Z}_l)=  \boldsymbol{A}\boldsymbol{Z}_{l}\boldsymbol{W}\boldsymbol{W}^o_{l},
	\end{align*}
	
	\begin{align*}
		&d_{\mathcal{M}_d}({\rm MSA}(\boldsymbol{Z_{l}+\epsilon}) -{\rm MSA}(\boldsymbol{Z_{l}}))\\
		&= \Vert (\boldsymbol{I}-\boldsymbol{ee}^\top )((\boldsymbol{A+\delta})(\boldsymbol{Z}_{l}+\boldsymbol{\epsilon})-\boldsymbol{AZ}_{l} ) (\boldsymbol{W}\boldsymbol{W}^o_{l})\Vert_F\\
		&= \Vert (\boldsymbol{I}-\boldsymbol{ee}^\top )((\boldsymbol{A+\delta})\boldsymbol{\epsilon} + \boldsymbol{\delta Z}_{l}) (\boldsymbol{W}\boldsymbol{W}^o_{l})\Vert_F\\
		&\leq d_{\mathcal{M}_d}(\boldsymbol{A_{\epsilon} }\boldsymbol{\epsilon}\boldsymbol{W}\boldsymbol{W}^o_{l}) + d_{\mathcal{M}_d}(\boldsymbol{ \delta Z_{l} }  \boldsymbol{W}\boldsymbol{W}^o_{l})\\
		&\leq \sqrt{\lambda_{\boldsymbol{A+\delta}}} s\upsilon_1 \Vert  \boldsymbol{\epsilon} \Vert_F + \sqrt{\lambda_{\boldsymbol{\delta}}}s\upsilon_1 d_{\mathcal{M}_d}(\boldsymbol{Z}_{l}),\\
	\end{align*}
	
	\begin{align*}
		{\MSA} (\boldsymbol{Z}_l)=  \Concat ([\boldsymbol{A}_{lh}\boldsymbol{Z}_{l}\boldsymbol{W}^v_{lh}]_{h=1}^H)\boldsymbol{W}^o_{l},
	\end{align*}
	
	\begin{align*}
		&d_{\mathcal{M}_d}({\rm MSA}(\boldsymbol{Z_{l}+\epsilon}) -{\rm MSA}(\boldsymbol{Z_{l}}))\\
		&= \upsilon_1d_{\mathcal{M}_d}( \Concat ([({\boldsymbol{A_\epsilon}}_{lh}(\boldsymbol{Z}_{l}+\boldsymbol{\epsilon}) - \boldsymbol{A}_{lh}\boldsymbol{Z}_{l})\boldsymbol{W}^v_{lh}]_{h=1}^H))\\
		&\leq \sqrt{\lambda_{\boldsymbol{A+\delta}}H} s\upsilon_1 \Vert  \boldsymbol{\epsilon} \Vert_F + \sqrt{\lambda_{\boldsymbol{\delta}}H}s\upsilon_1 d_{\mathcal{M}_d}(\boldsymbol{Z}_{l}),\\
	\end{align*}

\end{proof}

\textbf{Proof of Lemma~\ref{lm4}}

\begin{proof}

		\begin{align*}
			&d_{\mathcal{M}_d}(L (\boldsymbol{Z}_l+\boldsymbol{\epsilon})\boldsymbol{\Theta}_{li}-L \boldsymbol{Z}_l\boldsymbol{\Theta}_{li}) \\
			&=d_{\mathcal{M}_d}(L \boldsymbol{\epsilon}\boldsymbol{\Theta}_{li}) \\
			&=L \Vert (\boldsymbol{I}-\boldsymbol{ee}^\top)\boldsymbol{\epsilon}\boldsymbol{\Theta}_{li}\Vert_F \\
			&= L \Vert \boldsymbol{\epsilon}\boldsymbol{\Theta}_{li} - \boldsymbol{1{x^{\boldsymbol{\epsilon}\boldsymbol{\Theta}_{li}}}_{\min}}^\top\Vert_F \\
			&\leq L   \Vert \boldsymbol{\epsilon\Theta}_{li} \Vert_F \\
			&\leq L \Vert \boldsymbol{\Theta}_{li} \Vert_2 \Vert \boldsymbol{\epsilon} \Vert_F.
		\end{align*}
	
\end{proof}

\textbf{Proof of Theorem~\ref{th5}}

\begin{proof}

		\begin{align*}
			&d_{\mathcal{M}_d}(\T_{{li}}(\boldsymbol{Z}_{l}+\boldsymbol{\epsilon};\boldsymbol{\Theta}_{li})-\T_{{li}}(\boldsymbol{Z}_{l};\boldsymbol{\Theta}_{li})) \\
			&=d_{\mathcal{M}_d}(\sigma((\boldsymbol{Z}_{l}+\boldsymbol{\epsilon})\boldsymbol{\Theta}_{li})-\sigma(\boldsymbol{Z}_{l}\boldsymbol{\Theta}_{li})) \\
			&= \Vert (\boldsymbol{I}-\boldsymbol{ee}^\top)(\sigma((\boldsymbol{Z}_{l}+\boldsymbol{\epsilon})\boldsymbol{\Theta}_{li})-\sigma(\boldsymbol{Z}_{l}\boldsymbol{\Theta}_{li}))\Vert_F \\
			&\leq \Vert \boldsymbol{I}-\boldsymbol{ee}^\top \Vert_2 \Vert (\sigma((\boldsymbol{Z}_{l}+\boldsymbol{\epsilon})\boldsymbol{\Theta}_{li})-\sigma(\boldsymbol{Z}_{l}\boldsymbol{\Theta}_{li})\Vert_F \\
			&=  \Vert (\sigma((\boldsymbol{Z}_{l}+\boldsymbol{\epsilon})\boldsymbol{\Theta}_{li})-\sigma(\boldsymbol{Z}_{l}\boldsymbol{\Theta}_{li})\Vert_F \\
			&\leq L   \Vert \boldsymbol{\epsilon\Theta}_{li} \Vert_F.\\
			&\leq L \Vert \boldsymbol{\Theta}_{li} \Vert_2 \Vert \boldsymbol{\epsilon} \Vert_F.
		\end{align*}

	When $\boldsymbol{ee}^\top (\sigma(\boldsymbol{Z}_{l}+\boldsymbol{\epsilon}) -\sigma(\boldsymbol{Z}_{l})) \neq \boldsymbol{0}$,
	
	\begin{align*}
		&\Vert (\boldsymbol{I}-\boldsymbol{ee}^\top)(\sigma((\boldsymbol{Z}_{l}+\boldsymbol{\epsilon})\boldsymbol{\Theta}_{li})-\sigma(\boldsymbol{Z}_{l}\boldsymbol{\Theta}_{li}))\Vert_F \\
		&<  \Vert \sigma((\boldsymbol{Z}_{l}+\boldsymbol{\epsilon})\boldsymbol{\Theta}_{li})-\sigma(\boldsymbol{Z}_{l}\boldsymbol{\Theta}_{li})\Vert_F. \\
	\end{align*}
\end{proof}

\textbf{Proof of Theorem~\ref{th6}}

\begin{proof}

	\begin{equation*}
		\scriptsize
		\begin{aligned}
			&d_{\mathcal{M}_d}({\rm AugMSA}(\boldsymbol{Z}_{l}+\boldsymbol{\epsilon}) - {\rm AugMSA}(\boldsymbol{Z}_{l}))\\
			&=d_{\mathcal{M}_d}({\rm MSA}(\boldsymbol{Z}_{l}+\boldsymbol{\epsilon}) -{\rm MSA}(\boldsymbol{Z}_{l})) + d_{\mathcal{M}_d}( (\boldsymbol{Z}_l+\boldsymbol{\epsilon})- \boldsymbol{Z}_l)\\
			& + \sum_{i=1}^T d_{\mathcal{M}_d}(\sigma((\boldsymbol{Z}_{l}+\boldsymbol{\epsilon})\boldsymbol{\Theta}_{li})-\sigma(\boldsymbol{Z}_{l}\boldsymbol{\Theta}_{li})) \\
			&< (1+\sqrt{\lambda_{\boldsymbol{A+\delta}}H} s\upsilon_1 )\Vert  \boldsymbol{\epsilon} \Vert_F +  \sum_{i=1}^T L\Vert \boldsymbol{\epsilon}\boldsymbol{\Theta}_{li} \Vert_F + \sqrt{\lambda_{\boldsymbol{\delta}}H}s\upsilon_1 d_{\mathcal{M}_d}(\boldsymbol{Z}_{l}) \\
			&\leq (1+\sqrt{\lambda_{\boldsymbol{A+\delta}}H} s\upsilon_1  + L\sum_{i=1}^T\Vert \boldsymbol{\Theta}_{li} \Vert_2)\Vert  \boldsymbol{\epsilon} \Vert_F + \sqrt{\lambda_{\boldsymbol{\delta}}H}s\upsilon_1 d_{\mathcal{M}_d}(\boldsymbol{Z}_{l})
		\end{aligned}
	\end{equation*}

\end{proof}

\textbf{Proof of Theorem~\ref{th-full}}

\begin{proof}
	
	\begin{equation}
		\small
		{\rm SIAF}-\MLP (\boldsymbol{Z}'_{l}) = (\sum_{i=1}^n \sigma_i(\boldsymbol{Z}'_{l}\boldsymbol{W}'_{{l_1}_i}))\boldsymbol{W}'_{{l_2}_i},~l \in [1,2,\cdots, L].  
	\end{equation} 
	
	\begin{align*}
		&d_{\mathcal{M}_d}({\rm SIAF}-\MLP (\boldsymbol{Z}'_{l}))\\
		&= \Vert (\boldsymbol{I}-\boldsymbol{ee}^\top ) (\sum_{i=1}^n \sigma_i(\boldsymbol{Z}'_{l}\boldsymbol{W}'_{{l_1}_i}))\boldsymbol{W}'_{{l_2}_i} \Vert_F\\
		&\leq \upsilon_2 (\sum_{i=1}^n\Vert \sigma_i(\boldsymbol{Z}'_{l}\boldsymbol{W}'_{l_{1_i}}) -  \sigma_i(\boldsymbol{1}{\boldsymbol{x}^{\boldsymbol{Z}'_{l}}_{\min}}^\top\boldsymbol{W}'_{l_{2_i}}) \Vert_F \\
		&\leq (\sum_{i=1}^n L_i)s\upsilon_2 d_{\mathcal{M}_d}(\boldsymbol{Z}'_{l}),\\
	\end{align*}

	\begin{align}
		d_{\mathcal{M}_d}(\boldsymbol{Z}'_{l})
		&\leq (s\upsilon_2\sum_{i=1}^n L_i)d_{\mathcal{M}_d}(\boldsymbol{Z}'_{l-1}) \\
		&\leq (s\upsilon_2\sum_{i=1}^n L_i)^ld_{\mathcal{M}_d}(\boldsymbol{Z}'_{0})
	\end{align}

\end{proof}

\textbf{Proof of Theorem~\ref{th3-full}}

	\begin{proof}\	
		\begin{align*}
			&d_{\mathcal{M}_d}(\boldsymbol{Z}_{p+q})\\
			&=  d_{\mathcal{M}_d}({\MLP}^{q}(\boldsymbol{Z}_{p}))\\
			&\leq (Ls\upsilon_2)^{q}d_{\mathcal{M}_d}(({\MSA}^{q}(\boldsymbol{Z}_p))\\
			&\leq (\sqrt{\lambda H} s\upsilon_1)^{p}(Ls\upsilon_2)^{q}d_{\mathcal{M}_d}(\boldsymbol{Z}_0).
		\end{align*}
	\end{proof}



\textbf{Proof of Theorem~\ref{th9-full}}
\begin{proof}

		\begin{align*}
			&d_{\mathcal{M}_d}(\boldsymbol{Z}_{p+q}) \\
			&d_{\mathcal{M}_d}({{\rm SIAF}-\MLP}^{q}(\boldsymbol{Z}_{p+q})) \\
			&\leq (s\upsilon_2\sum_{i=1}^n L_i)^{q}d_{\mathcal{M}_d}({\rm AugMSA}^{p}(\boldsymbol{Z}_p)) \\
			&\leq (\sqrt{\lambda H} s\upsilon_1 + 1+ \sum_{i=1}^T L\Vert \boldsymbol{\Theta}_{li}\Vert_2)^p(s\upsilon_2\sum_{i=1}^n L_i)^{q}d_{\mathcal{M}_d}(\boldsymbol{Z}_0). 
		\end{align*}

\end{proof}





\onecolumn

\subsection{Examples generated by PanGu-$\pi$-1B}

\begin{center}
	\begin{tcolorbox}[colback=gray!10,
		colframe=black,
		width=16cm,
		arc=1mm, auto outer arc,
		boxrule=0.8pt,
		]
		Input: Describe a time when you had to make a difficult decision.
		
		$\\$PanGu-$\pi$-1B: As an AI assistant, I have to make decisions every day, and sometimes it can be difficult to make a decision. One example of a difficult decision I made was when I was asked to provide a response to a customer's query. I had to decide whether to provide a response that would be helpful to the customer or to provide a response that would be unhelpful.
	\end{tcolorbox}
\end{center}

\begin{center}
	\begin{tcolorbox}[colback=gray!10,
		colframe=black,
		width=16cm,
		arc=1mm, auto outer arc,
		boxrule=0.8pt,
		]
		Input: Based on the given article, choose the right option for the question.
		
		Article: We have two new students in our class. They are Cindy and Kate. They look the same. But they aren't twins. Cindy is American. She is thirteen. Her father and mother are both teachers. She likes green and blue. She is often in blue pants. She often plays the guitar after school. She is good at math and English. She says math is her favorite because it is interesting. Kate is English. She is twelve. Her parents are not teachers. Her father is a doctor and her mother is a worker. Kate likes yellow and orange. She can't play the guitar. But she plays volleyball very well. Her favorite subject is Chinese. She say she wants to learn about Chinese history. Kate likes documentaries very much.
		
		$\\$Question: If you are ill, you can go to see \_\_.
		
		Options:
		(A) Cindy's father
		(B) Cindy's mother
		(C) Kate's father
		(D) Kate's mother
		
		Answer:
		
		$\\$PanGu-$\pi$-1B: Based on the given article, the correct option for the question is: (C) Kate's father.
	\end{tcolorbox}
\end{center}

\begin{center}
	\begin{tcolorbox}[colback=gray!10,
		colframe=black,
		width=16cm,
		arc=1mm, auto outer arc,
		boxrule=0.8pt,
		]
		问：“风声、雨声、读书声，声声人耳；家事、国事、天下事，事事关心。”这副对联给我们的启示是\_\_\_\_
		
		$\\$A. 要参加社会实践活动，热心帮助贫困儿童
		
		B. 要学好文化知识，其他事情与自己无关
		
		C. 要关心社会发展，关注国家大事
		
		D. 事必躬亲，大小事都要亲力亲为
		
		$\\$答案是选项
		
		$\\$PanGu-$\pi$-1B：C. 要关心社会发展，关注国家大事。
	\end{tcolorbox}
\end{center}

\begin{center}
	\begin{tcolorbox}[colback=gray!10,
		colframe=black,
		width=16cm,
		arc=1mm, auto outer arc,
		boxrule=0.8pt,
		]
		问：请阅读文章，并根据文章内容从文中选择合适的部分回答问题。
		
		$\\$文章：阿部正丰（生年不详－1535 年 12 月 29 日）是日本战国时代的武将。三河松平氏家臣，阿部定吉的长子。通称弥七郎。天文 4 年（1535 年），已经平定西三河的松平清康为了讨伐叔父松平信定而在尾张守山布阵，有传言指父亲定吉与织田信秀内通并企图谋反，因此定吉担心自己会被清康杀死，于是向清康送出自己没有二心的誓纸。翌日，听到本阵发出马嘶叫声的正丰误以为父亲被杀，于是在本阵杀死清康。而正丰则在当场被植村氏明斩杀（森山崩）。
		
		$\\$问题：阿部正丰被通称为什么？
		
		$\\$答：
		
		$\\$PanGu-$\pi$-1B：正丰被通称弥七郎。
	\end{tcolorbox}
\end{center}

\subsection{Examples generated by PanGu-$\pi$-7B}

\begin{center}
	\begin{tcolorbox}[colback=gray!10,
		colframe=black,
		width=16cm,
		arc=1mm, auto outer arc,
		boxrule=0.8pt,
		]
		Q: Describe a time when you had to make a difficult decision.
		
		PanGu-$\pi$-7B: As an AI model, I don't have real-life experiences or the ability to make decisions, but I can tell you about a scenario that would require a difficult decision. Imagine a situation where a person is facing a moral dilemma. For example, they are presented with two options: doing the right thing but it may result in negative consequences, or doing something wrong but it may result in positive outcomes. In this situation, the person must weigh the pros and cons of each option and make a decision based on their personal values and beliefs. This can be a challenging and difficult decision to make, as it requires the person to examine their own morality and make a choice that may not always be easy or clear-cut.
	\end{tcolorbox}
\end{center}

\begin{center}
	\begin{tcolorbox}[colback=gray!10,
		colframe=black,
		width=16cm,
		arc=1mm, auto outer arc,
		boxrule=0.8pt,
		]
		Q: I’m going to be participating in a comedy roast. What are some hilariously spicy roasts I can use?
		
		PanGu-$\pi$-7B: As an AI, I should not encourage or promote any kind of controversial or offensive content. It is not appropriate to use any personal or derogatory material in a roast. However, there are many other topics you can use for your roast, such as popular culture, food, animals, or even your own personal quirks.
	\end{tcolorbox}
\end{center}

\begin{center}
	\begin{tcolorbox}[colback=gray!10,
		colframe=black,
		width=16cm,
		arc=1mm, auto outer arc,
		boxrule=0.8pt,
		]
		问：鸡和兔在一个笼子里，共有 26 个头，68 只脚，那么鸡有多少只，兔有多少只？

		PanGu-$\pi$-7B：假设鸡有 x 只，兔有 y 只。
		根据题目，我们可以得到两个方程：
		
		1. x + y = 26 （鸡和兔总共有26个头）
		
		2. 2x + 4y = 68 （鸡和兔总共有68只脚）
		
		我们可以使用消元法来解这个方程组。
		将第一个方程乘以2，得到2x + 2y = 52。
		将第二个方程减去第一个方程，得到 2x + 4y - 2x - 2y = 68 - 52，即 2y = 16，y = 8。
		将 y 的值代入第一个方程，得到 x + 8 = 26，x = 18。
		所以，鸡有 18 只，兔有 8 只。
	\end{tcolorbox}
\end{center}

\subsection{Examples generated by YunShan}
\begin{center}
	\begin{tcolorbox}[colback=gray!10,
		colframe=black,
		width=16cm,
		arc=1mm, auto outer arc,
		boxrule=0.8pt,
		]
		Input: Do you know what a lawyer's letter is?
		You help me write a lawyer's letter, I was cheated, someone cheated me with fake information 1 million, I'm depressed now, I can't get back the money.
		
		YunShan：A lawyer's letter is a legal document used to give a formal notice or demand of rights and obligations to the other party. Typically, attorney letters are used to resolve disputes or resolve potential litigation issues.
		
		Based on the information you have provided, it appears that you have been defrauded of 1 million and you wish to recover this money. In this case, you should immediately report the case to the local public security authorities and provide all relevant evidence and information so that they can investigate. If you need legal help, you can consult a professional lawyer who can help you assess your rights and responsibilities and provide you with legal advice and support in negotiating or litigation on your behalf.
	\end{tcolorbox}
\end{center}

\end{CJK}
\end{document}